\documentclass[twoside]{article}

\usepackage[preprint]{aistats2027}
\usepackage[round]{natbib}

\usepackage{bm} 
\usepackage{xcolor}
\usepackage{colortbl}
\usepackage{amsmath, amssymb, amsthm, epsfig, url, array}
\usepackage{mathtools} 
\usepackage{cleveref}

\theoremstyle{plain}
\newtheorem{theorem}{Theorem}[section]

\newtheorem{proposition}[theorem]{Proposition}
\newtheorem{lemma}[theorem]{Lemma}
\newtheorem{corollary}[theorem]{Corollary}
\theoremstyle{definition}
\newtheorem{definition}[theorem]{Definition}
\newtheorem{assumption}[theorem]{Assumption}

\Crefname{Theorem}{Theorem}{theorems}
\Crefname{Lemma}{Lemma}{lemmas}
\Crefname{Definition}{Definition}{Definitions}
\Crefname{Assumption}{Assumption}{Assumptions}
\Crefname{Corollary}{Cororally}{Corollaries}
\Crefname{Property}{Property}{Properties} 
\usepackage{threeparttable}
\usepackage{makecell, caption, booktabs, multirow, siunitx}

\theoremstyle{plain}
 
\Crefname{Constraint}{Constraint}{Constraints} 

\usepackage{algorithm}
\usepackage[noend]{algorithmic}

\usepackage{array}
\usepackage{xurl}

\begin{document}

%

%

\twocolumn[
\aistatstitle{Interpretable Graph Partitioning for Spatial Prediction}
\aistatsauthor{Yuta Shikuri}
\aistatsaddress{ 
The Graduate University for Advanced Studies, Tokyo, Japan \\
Tokio Marine Holdings, Inc., Tokyo, Japan \\
shikuriyuta@gmail.com
} 
]

\begin{abstract}
Spatial prediction estimates unobserved values from spatially distributed observations. 
Although flexible predictive models are increasingly available, their practical use remains limited when interpretability is essential. 
In this study, we formulate a graph partitioning problem to construct a global surrogate model that approximates black-box predictions with simpler representations. 
Specifically, we select connected subgraphs to minimize within-cluster residuals in vertex-level predictions. 
Although this formulation can be viewed as a variant of spatial regionalization, it operates on model-based predictions at individual locations rather than predefined areal units. 
With many locations, this point-level formulation may produce artificial connections between regions, reducing interpretability. 
We therefore introduce a constraint that excludes such connections while retaining flexible partition boundaries. 
We theoretically analyze how this constraint affects the resulting partition. 
We also develop a computational procedure for adjusting partition boundaries. 
Experimental results show that the proposed method yields spatially coherent partitions with flexible boundaries. 
\end{abstract} 

\begin{figure*}[t]
\begin{center}
    \includegraphics[width=\linewidth]{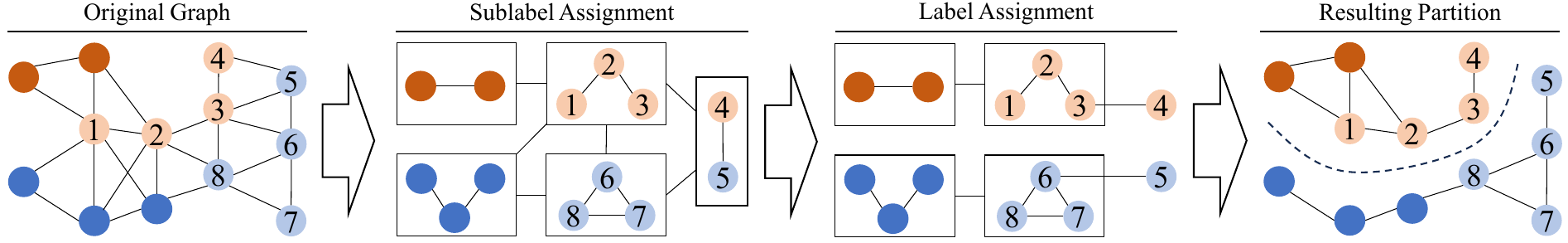}
\end{center}
\caption{
Overview of the proposed approach. 
Vertex colors represent the predicted values at individual locations. 
The locations are partitioned into spatially connected regions to approximate the predictions subject to a constraint defined using sublabels. 
}
\label{overview}
\end{figure*}

\section{Introduction} 
Spatial prediction aims to estimate unobserved values at new locations using observed data and their spatial dependence \citep{Diggle}. 
A wide range of methods has been developed for this task, among which Gaussian process regression \citep{Carl} has played a central role as a flexible Bayesian framework for modeling spatial dependence. 
Extensions of Gaussian process regression \citep{Chung,Yusuke,Shikuri} can further handle settings in which only aggregated or summarized observations are available, while enabling predictions at individual locations. 
In parallel, neural-network-based spatial predictors \citep{Wanfang,Gabriel} have been increasingly used for spatial prediction, allowing rich covariates and complex nonlinear relationships to be incorporated into predictive models.

Despite their expressive power, modern spatial prediction models are often difficult to deploy in domains where interpretability is essential. 
Although black-box models can provide fine-grained predictions for individual locations, many real-world workflows are organized around region-based management for planning, monitoring, or intervention. 
In such settings, decisions are not made independently for individual locations but are implemented through regional units that must be interpretable and operationally feasible. 
This creates a mismatch between the spatial granularity of model predictions and that of practical decision-making. 
Therefore, these predictions need to be summarized in a form that is interpretable at the regional level and consistent with operational decision rules. 

Interpretable machine learning \citep{Molnar} provides a useful framework for spatial prediction tasks. 
Surrogate models approximate black-box predictors using simpler and more interpretable representations, thereby facilitating the understanding, auditing, and debugging of predictive systems. 
These models are commonly classified as either local or global, depending on the scope of the approximation. 
Local surrogates \citep{Sameer,Scott} approximate a black-box model locally around individual inputs. 
In contrast, global surrogates approximate the overall predictive behavior of a black-box model using an interpretable structure, such as a decision tree \citep{Breiman}. 

Building on the idea of global surrogate modeling, we consider regional representations that approximate black-box spatial predictions by clustering individual locations into spatially contiguous regions. 
We formulate this task as a graph partitioning problem \citep{Ilya}, where vertices represent individual locations and edges encode spatial proximity. 
We then seek connected subgraphs with homogeneous predictions by minimizing within-cluster prediction errors. 
This formulation is closely related to conventional spatial regionalization \citep{Duque}. 
However, it differs in that regions are constructed from model-based predictions at individual locations, rather than from observed attributes attached to predefined areal units. 
This distinction can reveal complex spatial structures that do not align with areal boundaries, such as coastal areas exposed to substantial windstorm damage. 
At the same time, a standard connectivity constraint alone may permit artificial bridge-like connections between spatially separated regions. 

In this study, we introduce a constraint to rule out artificial bridge-like connections while allowing flexible partition boundaries. 
We analyze how optimal partitions with this constraint differ from those obtained by graph partitioning over preconstructed spatial groups. 
We also develop an algorithm to adjust the boundaries of these groups by adapting the automatic zoning procedure (AZP) \citep{Openshaw}. 
Experimental results show that our approach produces spatial partitions that avoid artificial bridge-like connections. 
\Cref{overview} provides an overview of the proposed method. 
\Cref{Description} summarizes the notation. 
\Cref{Proofs} presents the proofs of the theoretical results.

\section{Spatial Regionalization}
\label{Preliminary} 
In spatial regionalization, a graph $G \coloneqq (V,E)$ represents the contiguity relationships among spatial units, where the vertices $V \coloneqq \{1, \dotsc, n\}$ correspond to the spatial units and the edges $E$ encode their adjacency. 
Given an observed response vector $\bm{y} \coloneqq (y_i)_{i \in V}$ with $y_i \in \mathbb{R}$, the goal is to partition the vertex set $V$ into at most $m$ disjoint connected subsets so as to minimize the $\ell_2$-norm of the within-cluster residuals. 
A partition of $V$ is represented by an assignment vector $\bm{\omega} \coloneqq (\omega_i)_{i \in V}$, where $\omega_i \in W \coloneqq \{1, \dotsc, m\}$ denotes the \emph{label} assigned to vertex $i$. 
The search space of this vector is defined as $\Omega_0 \coloneqq \{\bm{\omega} \in W^n : G[V_j(\bm{\omega})] \text{ is empty or connected for each } j \in W\}$, where $V_j(\bm{\omega}) \coloneqq \{i \in V : \omega_i = j\}$. 
Then a spatial regionalization problem can be formulated as 
\begin{align}
\label{regionalization}  
\min_{\bm{\omega} \in \Omega_0} \|\bm{y} - g(\bm{\omega}; \bm{y})\|_2, 
\end{align} 
where the cluster-wise projection $g(\bm{\omega}; \bm{y})$ is given by 
\begin{align}
g(\bm{\omega}; \bm{y}) \coloneqq \Bigl(\frac{1}{|V_{\omega_i}(\bm{\omega})|} \sum_{i^\prime \in V_{\omega_i}(\bm{\omega})} y_{i^\prime}\Bigr)_{i \in V}.  
\end{align} 
This problem is denoted by $\mathrm{GPP}(\Omega_0, \bm{y})$. 

The spatial units are typically predefined areal units, such as districts, municipalities, or grid cells. 
One reason is that observations at individual locations may contain substantial noise. 
While individual locations can in principle be regarded as spatial units, directly regionalizing such noisy point-level observations may produce connected components whose shapes are largely determined by local fluctuations. 
Thus, connectivity alone at the point level may not be sufficient to obtain regional partitions that are interpretable as meaningful spatial regions.

\section{Interpretable Partition} 
\label{Setting} 
Unlike conventional spatial regionalization, we use individual location points $(\bm{x}_i)_{i \in V}$ with $\bm{x}_i \in \mathcal{X}$, as spatial units. 
The corresponding prediction vector is defined as $\bm{f} \coloneqq (f(\bm{x}_i))_{i \in V}$, where $f : \mathcal{X} \to \mathbb{R}$ is a spatial predictor trained separately from the partitioning step. 
In this setting, we first consider the problem $\mathrm{GPP}(\Omega_0, \bm{f})$, which can be related to the minimization of the squared prediction risk of a piecewise-constant spatial surrogate; see \Cref{background}. 

\begin{figure}[t] 
\begin{center}
\includegraphics[width=\linewidth]{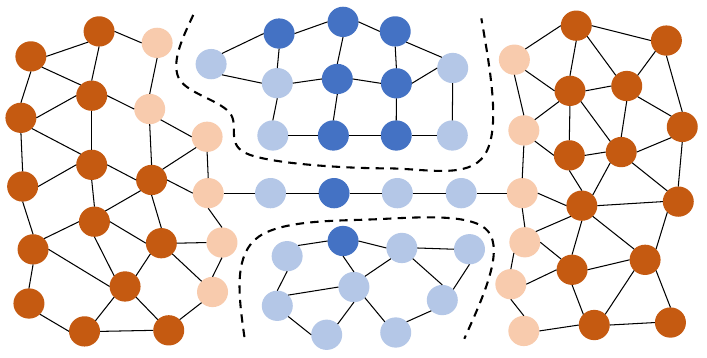} 
\end{center} 
\caption{
Example of a bridge-like connection. 
Two spatially separated regions of beige vertices are connected through intermediate vertices consisting of one blue vertex and three light-blue vertices. 
Such a connection may be introduced solely to decrease the objective value, but it does not reflect a spatially coherent regional structure. 
}
\label{ps} 
\end{figure}

Although $\mathrm{GPP}(\Omega_0, \bm{f})$ has the same mathematical structure as spatial regionalization, it uses the prediction vector $\bm{f}$ instead of the observed response vector $\bm{y}$. 
This distinction helps avoid uninterpretable partitions caused by noisy point-level observations. 
Nevertheless, imposing only the connectivity constraint $\bm{\omega} \in \Omega_0$ is still insufficient to ensure spatially coherent partitions of individual location points. 
For instance, as illustrated in \Cref{ps}, geographically separated regions with similar prediction levels may be assigned to the same partition if they are connected through a small number of intermediate vertices with substantially different prediction levels. 
Although such a bridge-like connection is mathematically feasible, it is difficult to interpret as a spatially coherent region with homogeneous prediction values. 

To construct spatial groups, we introduce an assignment vector $\bm{\sigma} \coloneqq (\sigma_i)_{i \in V}$, 
where $\sigma_i \in V_\circ \coloneqq \{1,\dotsc,l\}$ denotes the \emph{sublabel} assigned to vertex $i$, with $l \leq n$.  
For each $k \in V_\circ$, let $S_k \coloneqq \{i \in V : \sigma_i = k\}$. 
We require that $S_k \neq \emptyset$ and that $G[S_k]$ is connected. 
Given a label assignment vector $\bm{\omega}$, each sublabel group $k \in V_\circ$ is called a \emph{core group} if $|\{\omega_i : i \in S_k\}| = 1$; otherwise, it is called a \emph{boundary group}. 
We regard a connection between sublabel groups as \emph{bridge-like} if it is mediated solely by boundary groups. 
In this study, interpretability refers to representing predictions by a small number of connected spatial regions without artificial bridge-like connections. 

A naive way to exclude bridge-like connections is to prohibit boundary groups altogether by restricting the search space $\Omega_0$ to 
$\underline{\Omega} \coloneqq \{\bm{\omega} \in \Omega_0 : |\{\omega_i : i \in S_k\}| = 1 \text{ for each } k \in V_\circ\}$. 
This restriction also yields the following reformulation. 
\begin{proposition} 
\label{aggregation} 
The set of optimal label assignment vectors for $\mathrm{GPP}(\underline{\Omega}, \bm{f})$ coincides with that for $\mathrm{GPP}(\underline{\Omega}, \bm{f}_\sigma)$, 
where $\bm{f}_\sigma \coloneqq g(\bm{\sigma}; \bm{f})$. 
\end{proposition} 

For $S \subseteq V$, let $E_\circ(S) \coloneqq \{\{\sigma_i, \sigma_{i^\prime}\} : i, i^\prime \in S, \{i, i^\prime\} \in E, \sigma_i \neq \sigma_{i^\prime}\}$. 
By \Cref{aggregation}, $\mathrm{GPP}(\underline{\Omega}, \bm{f})$ can be reduced to a weighted graph partitioning problem on the aggregated graph $G_\circ \coloneqq (V_\circ, E_\circ(V))$. 
This formulation is computationally more tractable than $\mathrm{GPP}(\Omega_0, \bm{f})$. 
However, restricting the solution to $\underline{\Omega}$ may produce coarse partition boundaries that are constrained by the preconstructed sublabel groups. 

\begin{figure}[t] 
\begin{center}
\includegraphics[width=\linewidth]{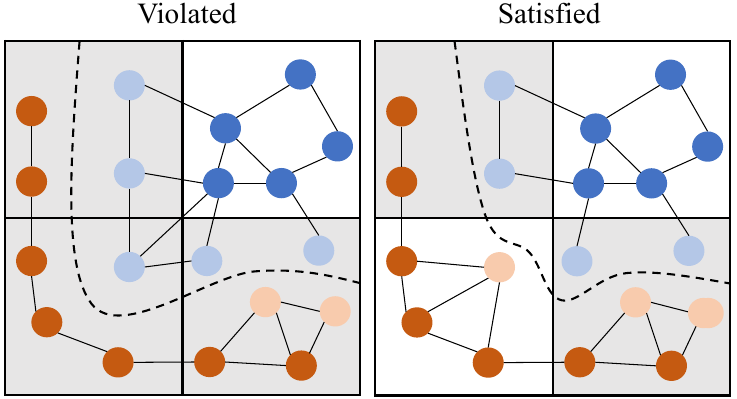} 
\end{center} 
\caption{
Illustration of \Cref{def_omega}. 
The left and right panels show two independent graph partitioning instances with $m = 2$. 
Each square represents a sublabel group. 
The dashed line indicates the boundary between the brown--beige partition $V_1(\bm{\omega})$ and the blue--sky-blue partition $V_2(\bm{\omega})$. 
White and gray squares represent core and boundary groups, respectively. 
In the left panel, let $S$ denote the bottom-left gray square. 
Since $G[V_1(\bm{\omega}) \setminus (S \cap V_1(\bm{\omega}))]$ is disconnected, this assignment violates the condition of \Cref{def_omega}. 
}
\label{example_omega}
\end{figure} 

To allow more flexible boundaries, we define the following set of label assignment vectors that imposes no restriction beyond excluding bridge-like connections. 
\begin{definition} 
\label{def_omega} 
Let $\Omega$ be the largest subset of $\Omega_0$ such that, for each $\bm{\omega} \in \Omega$ and each $j \in W$, 
$G[V_j(\bm{\omega}) \setminus \bigcup_{S \in \mathcal{S}} S]$ is empty or connected 
for every $\mathcal{S} \subseteq \mathcal{B}_j(\bm{\omega}) \coloneqq \{S_k \cap V_j(\bm{\omega}) : k \in V_\circ \text{ satisfying } S_k \not\subseteq V_j(\bm{\omega})\} \setminus \{\emptyset\}$. 
\end{definition} 
\Cref{example_omega} illustrates the role of the condition in \Cref{def_omega}. 
This condition is stated for every subcollection $\mathcal{S} \subseteq \mathcal{B}_j(\bm{\omega})$, rather than only for $\mathcal{S} = \mathcal{B}_j(\bm{\omega})$. 
Otherwise, a small subset of the vertices in a core group could be split off and treated as a boundary group. 
The condition in \Cref{def_omega} prevents such artificial boundary groups from occurring. 
With this definition, we analyze the partitions obtained from $\mathrm{GPP}(\Omega, \bm{f})$ and develop an approximate solution method. 

\begin{figure*}[t]
\begin{center}
\includegraphics[width=\linewidth]{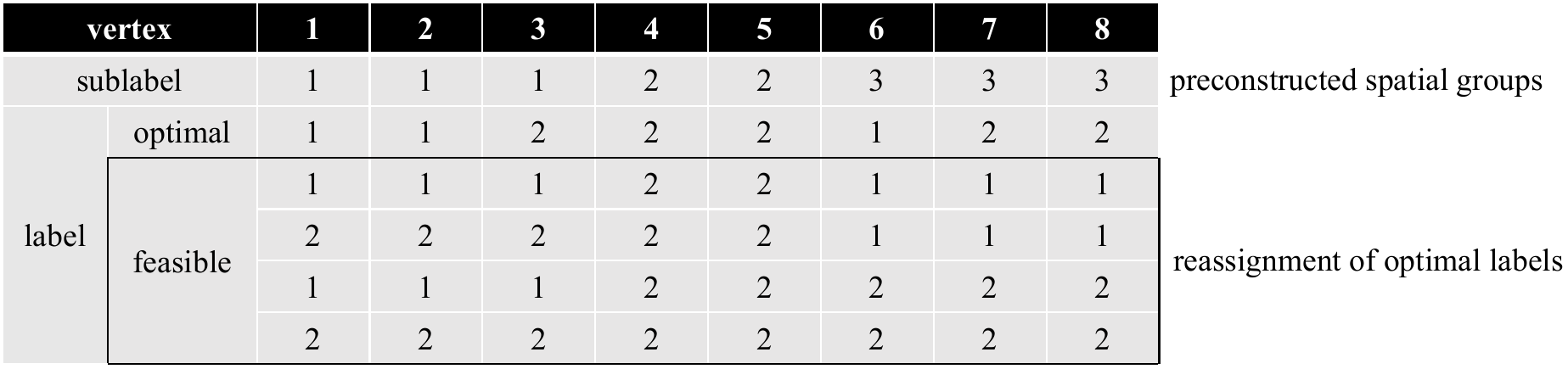}
\end{center}
\caption{ 
Example of label reassignment.  
Let $n = 8, m = 2$, and $l = 3$. 
This table illustrates how optimal labels are reassigned for the first and third sublabel groups. 
For the second sublabel group, no reassignment is performed because its assignment in the optimal partition has only a single label. 
} 
\label{replacement} 
\end{figure*}

\section{Boundary Flexibility}  
\label{Analysis} 
The constraint $\bm{\omega} \in \Omega$ is central to our formulation. 
It allows more flexible partition boundaries than the stronger constraint $\bm{\omega} \in \underline{\Omega}$. 
In this section, we examine how $\mathrm{GPP}(\Omega, \bm{f})$ differs from $\mathrm{GPP}(\underline{\Omega}, \bm{f}_\sigma)$. 

\subsection{Label Reassignment} 
For the analysis of flexible partition boundaries, we consider the label reassignment operation illustrated in \Cref{replacement}. 
To facilitate this analysis, we first define the following set of label assignment vectors. 
\begin{definition} 
\label{def_omega1} 
Let $\overline{\Omega}$ be the largest subset of $W^n$ such that each $\bm{\omega} \in \overline{\Omega}$ satisfies 
\begin{align} 
(c_{\sigma_i})_{i \in V} \in \underline{\Omega}  
\end{align} 
for every choice $(c_k)_{k \in V_\circ}$ satisfying $c_k \in \{\omega_i : i \in S_k\}$ for all $k \in V_\circ$. 
\end{definition} 
\begin{proposition} 
\label{trivial_relation} 
$\underline{\Omega} \subseteq \overline{\Omega}$. 
\end{proposition} 

When $\bm{f}_\sigma$ provides a good piecewise-constant approximation to the individual predictions $\bm{f}$, 
imposing the constraint $\bm{\omega} \in \underline{\Omega}$ may not substantially increase the optimal value relative to imposing $\bm{\omega} \in \overline{\Omega}$. 
This is because, under the label reassignment operation, we can choose a label for each sublabel group so that the resulting objective value remains close to the optimal value over $\overline{\Omega}$. 
The following lemma formalizes this intuition in the idealized case where the piecewise-constant approximation is exact. 
\begin{lemma} 
\label{identical_solution} 
Every optimal label assignment vector for $\mathrm{GPP}(\underline{\Omega}, \bm{f}_\sigma)$ is also optimal for  
\begin{align} 
\label{GPP_semi}  
\min_{\bm{\omega} \in \overline{\Omega}} \|\bm{f}_\sigma - g(\bm{\omega}; \bm{f})\|_2. 
\end{align} 
\end{lemma} 

By \Cref{identical_solution}, in the case where $\bm{f} = \bm{f}_\sigma$, the search space of $\mathrm{GPP}(\overline{\Omega}, \bm{f})$ can be reduced from $\overline{\Omega}$ to $\underline{\Omega}$. 
However, even when $\bm{f}$ is well approximated by the piecewise-constant vector $\bm{f}_\sigma$, the two vectors rarely coincide exactly in practice. 
This raises the question of how well $\mathrm{GPP}(\underline{\Omega}, \bm{f}_\sigma)$ approximates $\mathrm{GPP}(\overline{\Omega}, \bm{f})$ in view of the discrepancy between $\bm{f}$ and $\bm{f}_\sigma$. 
The following theorem addresses this question. 
\begin{theorem} 
\label{guarantee_underline} 
Let $\underline{\bm{\omega}}$ and $\overline{\bm{\omega}}$ be optimal label assignment vectors for $\mathrm{GPP}(\underline{\Omega}, \bm{f}_\sigma)$ and $\mathrm{GPP}(\overline{\Omega}, \bm{f})$, respectively.  
Then 
\begin{align} 
0 \leq \epsilon(\underline{\bm{\omega}}, \overline{\bm{\omega}}) \leq \delta \coloneqq 2 \|\bm{f} - \bm{f}_\sigma\|_2, 
\end{align} 
where $\epsilon(\underline{\bm{\omega}}, \overline{\bm{\omega}}) \coloneqq \|\bm{f} - g(\underline{\bm{\omega}}; \bm{f})\|_2 - \|\bm{f} - g(\overline{\bm{\omega}}; \bm{f})\|_2$.  
\end{theorem} 
  
\Cref{guarantee_underline} establishes an upper bound on the objective-value gap $\epsilon(\underline{\bm{\omega}}, \overline{\bm{\omega}})$ between the alternative assignment $\underline{\bm{\omega}}$ and the optimal assignment $\overline{\bm{\omega}}$. 
Since this result is proved using the condition in \Cref{def_omega1}, it remains valid for any set nested between $\underline{\Omega}$ and $\overline{\Omega}$. 
The following lemma clarifies where $\Omega$ lies in this structure. 
\begin{definition} 
\label{lem_omega_def} 
Let $\Omega^\prime$ be the largest subset of $W^n$ such that each $\bm{\omega} \in \Omega^\prime$ satisfies  
\begin{align} 
(\omega_i \mathbf{1}_{\{\sigma_i \notin S_\circ\}} + c_{\sigma_i}\mathbf{1}_{\{\sigma_i \in S_\circ\}})_{i \in V} \in \Omega_0  
\end{align} 
for every $S_\circ \subseteq V_\circ$ and every choice $(c_k)_{k \in S_\circ}$ satisfying $c_k \in \{\omega_i : i \in S_k\}$ for all $k \in S_\circ$.  
\end{definition}
\begin{lemma} 
\label{lem_omega} 
$\Omega = \Omega^\prime$. 
\end{lemma} 

Reassigning label $j$ to another label for a subset $S_\circ \subseteq V_\circ$ corresponds to removing the union of a subcollection $\mathcal{S} \subseteq \mathcal{B}_j(\bm{\omega})$.  
As established in \Cref{lem_omega}, the set $\Omega$ admits an equivalent characterization in terms of such label reassignments. 
This characterization leads to the following inclusion relations. 
\begin{theorem} 
\label{lem_omega21} 
$\underline{\Omega} \subseteq \Omega \subseteq \overline{\Omega}$. 
\end{theorem} 

By \Cref{lem_omega21}, the result in \Cref{guarantee_underline} remains valid even when $\overline{\Omega}$ is replaced by $\Omega$. 
\begin{corollary} 
\label{guarantee0} 
Let $\bm{\omega}_*$ be an optimal label assignment vector for $\mathrm{GPP}(\Omega, \bm{f})$. 
Then 
\begin{align} 
0 \leq \epsilon(\underline{\bm{\omega}}, \bm{\omega}_*) \leq \delta. 
\end{align} 
\end{corollary} 

\Cref{guarantee0} suggests that the objective-value gap between $\mathrm{GPP}(\Omega, \bm{f})$ and $\mathrm{GPP}(\underline{\Omega}, \bm{f})$ may increase with $\delta$. 
A large $\delta$ can arise when the sublabel groups are too coarse to capture local spatial structure or when prediction values vary sharply across neighboring locations.

\subsection{Objective-Gap Bound} 
The key idea behind \Cref{guarantee0} is to replace the individual predictions $\bm{f}$ in $\mathrm{GPP}(\Omega, \bm{f})$ with the piecewise-constant approximation $\bm{f}_\sigma$. 
This replacement enables the label reassignment operation that reduces the search space from $\Omega$ to $\underline{\Omega}$. 
Although this reassignment treats all sublabel groups as potential boundary groups, 
the condition in \Cref{def_omega} restricts the possible patterns of $C_\circ(\bm{\omega}) \coloneqq \bigcup_{j \in W} C_j^\circ(\bm{\omega})$ and $B_\circ(\bm{\omega}) \coloneqq \bigcup_{j \in W} B_j^\circ(\bm{\omega})$, 
where $C_j^\circ(\bm{\omega}) \coloneqq \{\sigma_i : i \in C_j(\bm{\omega})\}$, 
$C_j(\bm{\omega}) \coloneqq \bigcup_{k \in V_\circ : S_k \subseteq V_j(\bm{\omega})} S_k$, 
and $B_j^\circ(\bm{\omega}) \coloneqq \{\sigma_i : i \in V_j(\bm{\omega}) \setminus C_j(\bm{\omega})\}$. 
To exploit this restriction, we first establish the following lemma, which can be derived in the same way as \Cref{identical_solution}. 
\begin{lemma} 
\label{identical_solution_p} 
For each $\bm{\omega} \in \Omega$, let $\Omega_p(\bm{\omega}) \coloneqq \{\bm{\omega}_p \in \Omega : C_j^\circ(\bm{\omega}) \subseteq C_j^\circ(\bm{\omega}_p) \text{ for each } j \in W\}$. 
Then every optimal label assignment vector for $\mathrm{GPP}(\underline{\Omega} \cap \Omega_p(\bm{\omega}), \bm{f}_\sigma)$ is also optimal for 
\begin{align} 
\label{GPP_semi_p} 
\min_{\bm{\omega}_p \in \Omega_p(\bm{\omega})} \|\bm{f}_\sigma - g(\bm{\omega}_p; \bm{f})\|_2. 
\end{align} 
\end{lemma} 

Using \Cref{identical_solution_p}, we derive a lower bound on the objective value of $\mathrm{GPP}(\Omega_p(\bm{\omega}), \bm{f})$ by replacing selected entries of $\bm{f}$ with the corresponding entries of $\bm{f}_\sigma$. 
\begin{theorem} 
\label{LB_sub} 
For $S_\circ \subseteq V_\circ$, let 
\begin{align} 
\bm{f}_p(S_\circ) \coloneqq ([\bm{f}]_i \mathbf{1}_{\{\sigma_i \notin S_\circ\}} + [\bm{f}_\sigma]_i \mathbf{1}_{\{\sigma_i \in S_\circ\}})_{i \in V}, \\  
\eta(S_\circ) \coloneqq \|\bm{f}_p(S_\circ) - g(\underline{\bm{\omega}}; \bm{f})\|_2 - \|\bm{f} - \bm{f}_p(S_\circ)\|_2. 
\end{align} 
Then, for each $\bm{\omega} \in \Omega$, 
\begin{align} 
\epsilon(\underline{\bm{\omega}}, \bm{\omega}) \leq \|\bm{f} - g(\underline{\bm{\omega}}; \bm{f})\|_2 - \eta(B_\circ(\bm{\omega})). 
\end{align} 
\end{theorem} 

Applying \Cref{LB_sub} for all $\bm{\omega} \in \Omega$, we obtain the following upper bound on $\epsilon(\underline{\bm{\omega}}, \bm{\omega}_*)$. 
\begin{corollary} 
\label{LB} 
Let $\mathcal{B}_\circ \coloneqq \{B_\circ(\bm{\omega}) : \bm{\omega} \in \Omega\}$. 
Then 
\begin{align} 
\epsilon(\underline{\bm{\omega}}, \bm{\omega}_*) \leq \zeta \coloneqq \|\bm{f} - g(\underline{\bm{\omega}}; \bm{f})\|_2 - \min_{S_\circ \in \mathcal{B}_\circ} \eta(S_\circ). 
\end{align} 
\end{corollary} 

The first term of $\eta(S_\circ)$ depends on the within-sublabel variation in the predictions for the candidate core groups, whereas the second term captures that for the remaining boundary groups. 
Despite these two components, $\eta(S_\circ)$ has the following monotonicity property. 
\begin{proposition} 
\label{monotonicity} 
For $S_\circ \subseteq V_\circ$, $\eta(S_\circ)$ is strictly decreasing in $\|\bm{f} - \bm{f}_p(S_\circ)\|_2$.
\end{proposition} 

By \Cref{monotonicity}, we can compute the upper bound $\zeta$ by identifying $S_\circ \in \mathcal{B}_\circ$ that maximizes $\|\bm{f} - \bm{f}_p(S_\circ)\|_2$.
This bound is no larger than $\delta$. 
\begin{proposition} 
\label{tight_result} 
The following inequalities hold: 
\begin{align} 
\zeta \leq 2 \max_{S_\circ \in \mathcal{B}_\circ} \|\bm{f} - \bm{f}_p(S_\circ)\|_2 \leq \delta. 
\end{align} 
\end{proposition} 

\Cref{tight_result} highlights the distinction between the fixed boundaries imposed by $\underline{\Omega}$ and the flexible boundaries allowed by $\Omega$.

\section{Boundary Adjustment} 
\label{Computation} 
In this section, we develop a computational scheme for approximately solving $\mathrm{GPP}(\Omega, \bm{f})$. 
The scheme consists of two steps: first solving $\mathrm{GPP}(\underline{\Omega}, \bm{f}_\sigma)$, and then adjusting the partition boundaries induced by $\underline{\bm{\omega}}$. 
The second step can be formulated as $\mathrm{GPP}(\widetilde{\Omega}, \bm{f})$, 
where $\widetilde{\Omega} \coloneqq \{\bm{\omega} \in \Omega : C_j^\circ(\bm{\omega}) \subseteq C_j^\circ(\underline{\bm{\omega}}) \text{ for each } j \in W\}$. 
We approximately solve this problem by starting from $\underline{\bm{\omega}}$ and greedily relabeling connected subgroups of boundary vertices to decrease the objective. 
This boundary adjustment resembles an AZP-type procedure, but the search space $\widetilde{\Omega}$ differs from that used in standard AZP algorithms. 
To facilitate feasibility checks, we establish the following characterization of $\Omega$. 
\begin{definition} 
\label{feasible_set_def} 
Let $\Omega^{\prime\prime}$ be the largest subset of $W^n$ such that, for each $\bm{\omega} \in \Omega^{\prime\prime}$ and each $j \in W$,  
\begin{itemize}
\item $G[C_j(\bm{\omega})]$ is empty or connected, 
\item if $C_j(\bm{\omega}) = \emptyset$, then $G[\bigcup_{S \in \mathcal{S}} S]$ is empty or connected for every $\mathcal{S} \subseteq \mathcal{B}_j(\bm{\omega})$,  
\item if $C_j(\bm{\omega}) \neq \emptyset$, then $G[S \cup C_j(\bm{\omega})]$ is connected for all $S \in \mathcal{B}_j(\bm{\omega})$. 
\end{itemize}
\end{definition}
\begin{lemma} 
\label{feasible_set} 
$\Omega = \Omega^{\prime\prime}$. 
\end{lemma} 
In the left panel of \Cref{example_omega}, the brown--beige subgraph violates the condition in \Cref{feasible_set_def} for $C_1(\bm{\omega}) = \emptyset$, whereas the blue--sky-blue subgraph satisfies the condition for $C_2(\bm{\omega}) \neq \emptyset$. 
This condition clarifies how to check feasibility when the boundary groups are given. 

The possible sets of boundary groups are restricted to the collection $\widetilde{\mathcal{B}}_\circ \coloneqq \{B_\circ(\bm{\omega}) : \bm{\omega} \in \widetilde{\Omega}\}$. 
To construct a larger collection of candidate sets containing $\widetilde{\mathcal{B}}_\circ$, we introduce a group-level condition corresponding to the vertex-level condition in \Cref{feasible_set_def}. 
\begin{definition} 
\label{Omega_plus}
Let $\Omega_+$ be the largest subset of $W^n$ such that, for each $\bm{\omega} \in \Omega_+$ and each $j \in W$,  
\begin{itemize}
\item $G_\circ[C_j^\circ(\bm{\omega})]$ is empty or connected,   
\item if $C_j^\circ(\bm{\omega}) = \emptyset$, then $G_\circ[B_j^\circ(\bm{\omega})]$ is empty or complete,   
\item if $C_j^\circ(\bm{\omega}) \neq \emptyset$, then $G_\circ[\{k\} \cup C_j^\circ(\bm{\omega})]$ is connected for all $k \in B_j^\circ(\bm{\omega})$. 
\end{itemize}
\end{definition} 
\begin{theorem} 
\label{feasible_set_core} 
$\widetilde{\mathcal{B}}_\circ \subseteq \widetilde{\mathcal{B}}_\circ^+ \coloneqq \{B_\circ(\bm{\omega}) : \bm{\omega} \in \widetilde{\Omega}_+\}$,  
where $\widetilde{\Omega}_+ \coloneqq \{\bm{\omega} \in \Omega_+ : C_j^\circ(\bm{\omega}) \subseteq C_j^\circ(\underline{\bm{\omega}}) \text{ for each } j \in W\}$. 
\end{theorem} 

Although \Cref{feasible_set_core} allows us to generate candidate boundary-group sets without specifying the labels assigned to their vertices, relaxing $\Omega$ to $\Omega_+$ may introduce additional candidates. 
The following assumption ensures that the two collections $\widetilde{\mathcal{B}}_\circ$ and $\widetilde{\mathcal{B}}_\circ^+$ coincide. 
\begin{assumption} 
\label{splittability} 
For every $\bm{\omega} \in \widetilde{\Omega}_+$, there exists a set $U \subseteq \bigcup_{k \in B_\circ(\bm{\omega})} S_k$ such that 
\begin{itemize} 
\item $E_\circ((V \setminus \bigcup_{k \in B_\circ(\bm{\omega})} S_k) \cup U) = E_\circ(V \setminus U) = E_\circ(V)$, 
\item both $G[U \cap S_k]$ and $G[(V \setminus U) \cap S_k]$ are nonempty and connected for all $k \in B_\circ(\bm{\omega})$. 
\end{itemize} 
\end{assumption} 
\begin{proposition} 
\label{BB_equal} 
Under \Cref{splittability}, $\widetilde{\mathcal{B}}_\circ = \widetilde{\mathcal{B}}_\circ^+$. 
\end{proposition} 
\Cref{splittability} is more plausible when each sublabel group is well connected and has multiple edges to each neighboring sublabel group. 
This is particularly relevant to our setting, where vertices correspond to individual locations rather than areal units. 

Minimizing the objective over feasible vertex labels for all candidate sets of boundary groups is computationally demanding. 
To avoid this exhaustive evaluation, we select a promising set of boundary groups using the upper bound provided in \Cref{LB_sub}. 
Specifically, we minimize $\eta(B_\circ(\bm{\omega}))$ over $\bm{\omega} \in \widetilde{\Omega}_+$. 
Although this criterion does not necessarily identify the optimal set of boundary groups, it favors candidate sets for which the bound permits a larger potential reduction in the objective. 
By \Cref{monotonicity}, this is equivalent to maximizing $\|\bm{f} - \bm{f}_p(S_\circ)\|_2$ over $S_\circ \in \widetilde{\mathcal{B}}_\circ^+$. 
We denote a maximizer by $\widetilde{B}_\circ$. 
The squared objective and the search space can be represented linearly using binary variables. 
Hence, the candidate set of boundary groups can be obtained by solving the mixed-integer linear program (MILP) in \Cref{MILP}. 

Given the selected boundary groups $\widetilde{B}_\circ$, we reassign the labels of vertices in $\widetilde{B} \coloneqq \{i \in V : \sigma_i \in \widetilde{B}_\circ\}$.  
For each $j \in W$, let $\widetilde{B}_j \coloneqq \{i \in \widetilde{B} : [\underline{\bm{\omega}}]_i = j\}$. 
If $\widetilde{B}_j \neq \emptyset$, we partition $\widetilde{B}_j$ into $\kappa_j \in \{1, \dotsc, |\widetilde{B}_j|\}$ connected subgroups; otherwise, we set $\kappa_j = 0$. 
All vertices within a subgroup are simultaneously relabeled to the same label. 
We then iteratively relabel these subgroups, choosing at each iteration the relabeling that yields the largest decrease in the objective while keeping the assignment in $\widetilde{\Omega}$. 
At each iteration, at most $(m - 1) \sum_{j \in W} \kappa_j$ relabelings are evaluated. 
The procedure terminates when no feasible relabeling decreases the objective. 
Note that \Cref{splittability} is not required for this relabeling procedure. 

\begin{table*}[t]
    \centering
    \caption{
    Comparison of partitioning performance across methods. 
    The mean and standard deviation of the error and running time over $30$ trials are reported. 
    The error is defined as $\|\bm{f} - g([1,\ldots,1]^\top; \bm{f})\|_2^{-1} \|\bm{f} - g(\bm{\omega}; \bm{f})\|_2$, where $\bm{\omega}$ denotes the resulting assignment vector. 
    The error is reported as a percentage. 
    The running time includes the common preprocessing required to construct the sublabel groups. 
    }
    \begin{threeparttable}[]
    \begin{tabular}{ccccccccc} 
    \toprule
    \multirow{2.5}{*}{Target} & \multirow{2.5}{*}{$m$} & \multirow{2.5}{*}{$l$} & \multicolumn{2}{c}{REDCAP} & \multicolumn{2}{c}{MIQCP} & \multicolumn{2}{c}{Ours} \\ 	
    \cmidrule(lr){4-5} \cmidrule(lr){6-7} \cmidrule(lr){8-9} & & & Error (\%) & Time [s] & Error (\%) & Time [s] & Error (\%) & Time [s] \\ 
    \midrule 
    \multirow{6}{*}{Price} & 2 & 15 & $79.6~(10.2)$ & $11~(1)$ & $77.5~(9.1)$ & $15~(1)$ & $77.3~(9.1)$ & $23~(10)$ \\
    & 2 & 30 & $79.6~(10.2)$ & $11~(1)$ & $76.6~(9.4)$ & $29~(5)$ & $76.5~(9.4)$ & $38~(16)$ \\
    & 3 & 15 & $66.6~(5.6)$ & $11~(0)$ & $65.1~(5.5)$ & $22~(4)$ & $64.6~(5.3)$ & $109~(133)$ \\
    & 3 & 30 & $66.6~(5.6)$ & $11~(0)$ & $64.1~(4.8)$ & $195~(99)$ & $63.9~(4.9)$ & $272~(123)$ \\
    & 4 & 15 & $57.6~(3.6)$ & $11~(1)$ & $56.8~(3.6)$ & $54~(12)$ & $56.4~(3.7)$ & $136~(73)$ \\
    & 4 & 30 & $57.6~(3.6)$ & $11~(1)$ & $55.6~(4.0)$ & $928~(371)$ & $55.4~(4.0)$ & $1085~(398)$ \\
    \midrule
    \multirow{6}{*}{Income} & 2 & 15 & $89.0~(1.0)$ & $11~(0)$ & $87.3~(2.6)$ & $14~(2)$ & $87.1~(2.6)$ & $34~(36)$ \\
    & 2 & 30 & $89.0~(1.0)$ & $11~(0)$ & $86.7~(2.9)$ & $29~(7)$ & $86.5~(3.0)$ & $50~(35)$ \\
    & 3 & 15 & $78.9~(2.3)$ & $11~(0)$ & $75.6~(5.0)$ & $19~(3)$ & $75.0~(5.3)$ & $110~(82)$ \\
    & 3 & 30 & $78.9~(2.3)$ & $11~(0)$ & $74.1~(5.2)$ & $221~(109)$ & $73.8~(5.4)$ & $295~(125)$ \\
    & 4 & 15 & $66.4~(4.2)$ & $11~(0)$ & $64.2~(3.7)$ & $43~(11)$ & $63.8~(3.7)$ & $141~(70)$ \\
    & 4 & 30 & $66.4~(4.2)$ & $11~(0)$ & $63.3~(2.9)$ & $753~(335)$ & $63.0~(3.0)$ & $867~(356)$ \\
    \midrule
    \multirow{6}{*}{Building} & 2 & 15 & $75.2~(12.9)$ & $11~(0)$ & $59.9~(1.8)$ & $14~(1)$ & $59.8~(2.0)$ & $38~(33)$ \\
    & 2 & 30 & $75.2~(12.9)$ & $11~(0)$ & $58.5~(0.8)$ & $25~(5)$ & $58.4~(0.8)$ & $32~(11)$ \\
    & 3 & 15 & $59.4~(8.3)$ & $11~(0)$ & $47.3~(3.3)$ & $18~(3)$ & $47.0~(3.3)$ & $90~(114)$ \\
    & 3 & 30 & $59.4~(8.3)$ & $11~(0)$ & $46.1~(2.6)$ & $155~(45)$ & $45.9~(2.8)$ & $261~(148)$ \\
    & 4 & 15 & $43.1~(1.8)$ & $11~(0)$ & $39.9~(2.0)$ & $31~(5)$ & $39.4~(2.0)$ & $193~(188)$ \\
    & 4 & 30 & $43.1~(1.8)$ & $11~(0)$ & $38.8~(2.0)$ & $435~(167)$ & $38.3~(1.9)$ & $705~(347)$ \\
    \midrule
    \multirow{6}{*}{Agriculture} & 2 & 15 & $83.6~(1.1)$ & $11~(0)$ & $83.3~(1.0)$ & $13~(1)$ & $82.5~(1.2)$ & $32~(14)$ \\
    & 2 & 30 & $83.6~(1.1)$ & $11~(0)$ & $82.6~(0.6)$ & $26~(5)$ & $82.4~(0.7)$ & $43~(18)$ \\
    & 3 & 15 & $74.3~(3.3)$ & $11~(0)$ & $72.4~(2.4)$ & $18~(3)$ & $72.0~(2.5)$ & $62~(42)$ \\
    & 3 & 30 & $74.3~(3.3)$ & $11~(0)$ & $71.1~(1.9)$ & $176~(48)$ & $70.9~(2.1)$ & $254~(108)$ \\
    & 4 & 15 & $68.8~(2.6)$ & $11~(1)$ & $67.2~(1.5)$ & $44~(16)$ & $66.6~(1.8)$ & $136~(68)$ \\
    & 4 & 30 & $68.8~(2.6)$ & $11~(1)$ & $66.0~(1.7)$ & $922~(326)$ & $65.7~(1.7)$ & $1042~(383)$ \\
    \bottomrule 
    \end{tabular}
    \end{threeparttable} 
    \label{performance1}
\end{table*} 

\begin{figure*}[t] 
\begin{center}
\includegraphics[width=\linewidth]{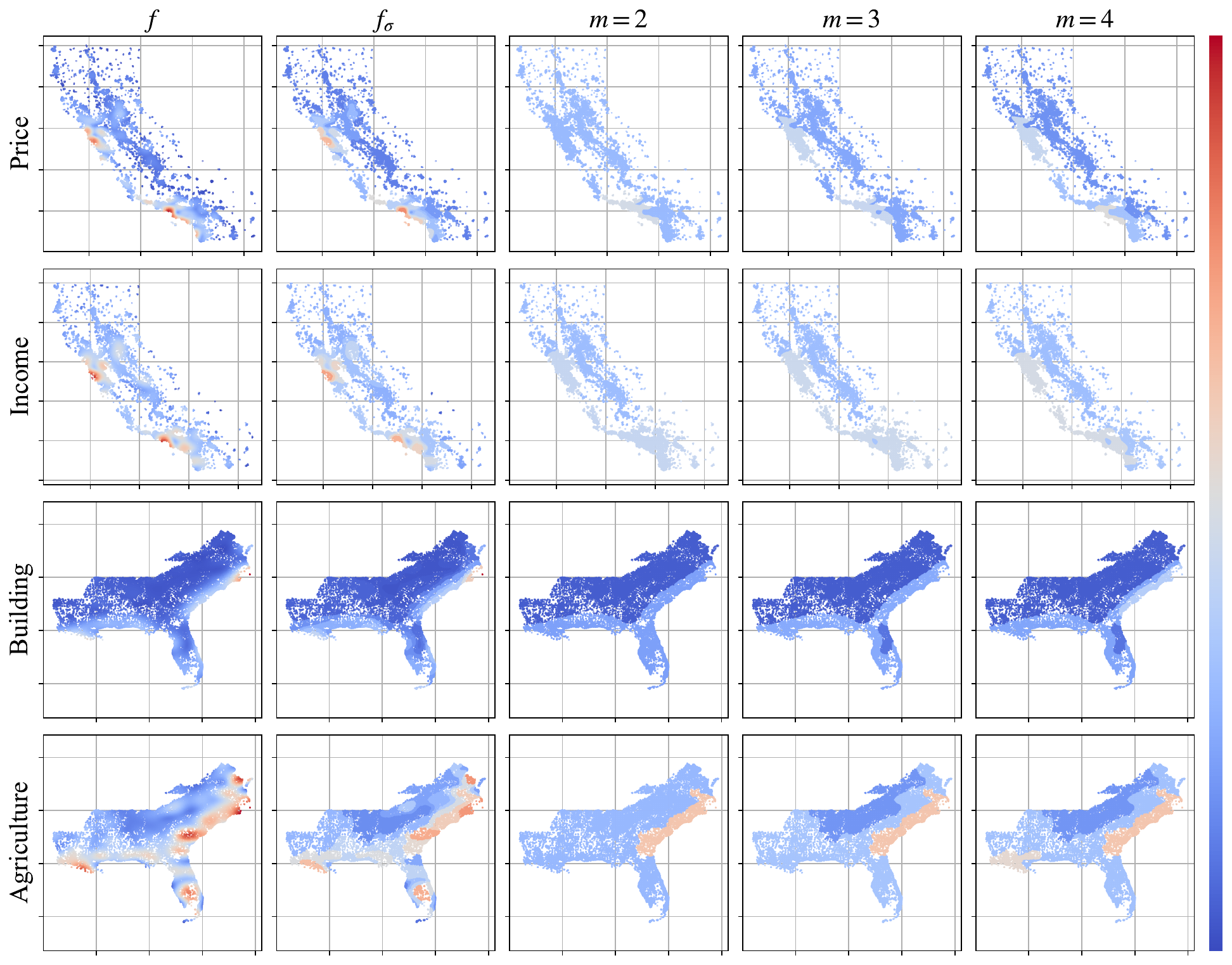}  
\end{center} 
\caption{
Spatial partitions produced by our approach with $l = 30$. 
For each target variable, we show the trial with the largest value of $\|\bm{f} - g([1,\ldots,1]^\top; \bm{f})\|_2$. 
The leftmost panel depicts the individual predictions $\bm{f}$. 
The second panel presents the piecewise-constant approximation $\bm{f}_\sigma$. 
The remaining panels give the cluster-wise projection $g(\bm{\omega}; \bm{f})$ for the resulting partitions. 
Colors indicate the levels of the predicted values. 
Colors are scaled separately for each target variable. 
}
\label{performance2}
\end{figure*}

\section{Numerical Study} 
\label{Experiment} 
In this section, we evaluate the proposed approach on real-world datasets. 
The data and software sources are provided in \Cref{source}. 

\textbf{Dataset.} 
We used two benchmark datasets. 
The first was the California Housing dataset, which contains $20{,}640$ samples. 
The target variable was either $\mathit{Price}$ or $\mathit{Income}$, linked to latitude-longitude pairs. 
The second was version 1.19.0 of the National Risk Index, from which we extracted $17{,}754$ samples corresponding to the states of Louisiana, Mississippi, Alabama, Florida, Georgia, South Carolina, North Carolina, and Virginia. 
These states form a contiguous region in the southeastern United States and provide substantial spatial variation in natural-hazard risk while maintaining a sufficiently large study area. 
The target variable was the expected annual loss rate for either $\mathit{Building}$ or $\mathit{Agriculture}$, associated with latitude-longitude pairs derived from 2024 Census Tract FIPS codes. 

\textbf{Spatial prediction.} 
We constructed the predictor $f$ using Gaussian process regression with variational inference \citep{Hensman}, employing an identity link function, a zero mean function, and $100$ inducing points. 
We initialized the inducing points using $k$-means clustering \citep{Vassilvitskii} with $10$ restarts. 
The covariance function was the product of a bias kernel and an ARD squared exponential kernel, plus a white-noise kernel. 
We used a Gaussian likelihood whose variance was treated as a hyperparameter. 
For each target variable, the hyperparameters were estimated using all available samples via L-BFGS \citep{Dong} with $10$ random restarts. 
For prediction, we generated $n = 10^5$ synthetic location points by resampling observations from each dataset with probabilities proportional to the population and adding isotropic perturbations with uniformly sampled directions and radii up to $0.05$ in the latitude-longitude coordinate system. 
This procedure was repeated over $30$ trials. 

\textbf{Graph partitioning.} 
For each trial, we solved the graph partitioning problem $\mathrm{GPP}(\Omega, \bm{f})$ to partition the $n = 10^5$ locations into $m \in \{2, 3, 4\}$ regions. 
A $1000$-nearest-neighbor graph weighted by Euclidean distances in the latitude-longitude coordinate system was first constructed. 
The graph $G$ was obtained by extracting a minimum spanning tree \citep{kruskal} from this graph and then augmented by adding an edge whenever either vertex was among the $10$ nearest neighbors of the other. 
We generated the preconstructed sublabel groups algorithmically rather than using existing regions, allowing us to consider multiple choices of $l$. 
Specifically, we used a Ward-type REDCAP algorithm \citep{Guo} to obtain $l \in \{15, 30\}$ sublabel groups because its sequence of local merges helps reduce bridge-like connections that can arise in global optimization. 
Starting from singleton groups, the algorithm repeatedly merges the pair of adjacent groups that yields the smallest increase in the within-group sum of squares of $\bm{f}$ until $l$ groups remain. 
The procedure in \Cref{Computation} was used to approximately solve $\mathrm{GPP}(\Omega, \bm{f})$. 
The initial label assignment vector $\underline{\bm{\omega}}$ was obtained by solving a mixed-integer quadratically constrained program (MIQCP),  
where the connectivity constraints were represented using a flow-based linear formulation \citep{Hojny}. 
The boundary groups $\widetilde{B}_\circ$ were selected by solving the MILP formulation. 
For each $j \in W$, the $\kappa_j = \min\{100, |\widetilde{B}_j|\}$ subgroups were constructed using the Ward-type REDCAP algorithm. 

\textbf{Baselines.} 
We used the Ward-type REDCAP algorithm as a baseline. 
Starting from the sublabel groups, we continued the merging procedure until $m$ groups remained. 
We also report the results obtained from the MIQCP formulation used to compute $\underline{\bm{\omega}}$. 
These baselines guarantee that the resulting partitions belong to $\underline{\Omega}$. 
Thus, they avoid bridge-like connections, provided that the sublabel groups are suitably chosen. 
However, their partition boundaries may be coarse. 
Comparison with these baselines therefore allows us to evaluate the effect of boundary adjustment. 

\textbf{Environment.} 
All experiments were conducted on a 64-bit Windows machine equipped with an Intel Core i9-9900K @ 3.60 GHz processor and 64 GB of RAM. 
The code was implemented in Python version 3.13.5. 
Gaussian process regression was performed using GPy version 1.14.2. 
The $k$-means clustering and nearest-neighbor graph construction were performed using scikit-learn version 1.6.1. 
Minimum spanning trees were constructed using SciPy version 1.15.3. 
The MILP and MIQCP problems were solved using SCIP version 10.0.2 through PySCIPOpt version 6.2.1. 
Unless otherwise specified, all software parameters were set to their default values. 

\textbf{Results.} 
\Cref{performance1} demonstrates that our approach achieves lower objective values than the baselines. 
Although the use of sublabels to avoid bridge-like connections restricts the search space, exact optimization of the resulting MIQCP consistently yielded lower objective values than REDCAP. 
Moreover, the boundary adjustment further reduced the objective by locally refining the coarse partition boundaries while preserving the global partition structure. 
One potential drawback is that the running time tends to increase for larger $m$ and $l$. 
\Cref{performance2} illustrates the spatially coherent partitions obtained under the proposed bridge-avoidance constraint. 
By treating individual locations as vertices rather than using predefined areal units, our problem formulation can capture coastal areas with higher predicted values. 
\Cref{bridge_like_connections} shows that further relaxing the bridge-avoidance constraint yields larger reductions in error but introduces bridge-like connections. 
\Cref{boundary_demo} shows how our approach modifies the boundaries of the MIQCP partitions.

\section{Conclusion} 
\label{Conclusion} 
In this study, we formulated a graph partitioning problem that summarizes the predictions of a black-box model into interpretable spatial regions. 
To obtain spatially coherent partitions of individual locations, we introduced a constraint that prevents bridge-like connections while allowing flexible partition boundaries. 
Theoretical results characterize how this constraint affects the resulting partitions. 
Experiments on real-world spatial datasets show that the proposed method produces spatially coherent partitions with lower approximation error than a standard spatial regionalization approach. 
Future work includes improving computational scalability for larger $m$ and $l$.

\newpage
\subsection*{AI Use Statement}

In this work, we used generative AI tools to assist with mathematical proof checking and to provide feedback on research methodology and implementation. 
Additionally, we used generative AI tools for code editing and debugging, and for drafting and editing the manuscript to improve clarity and readability. 
All AI-assisted mathematical arguments, code, and manuscript text were independently reviewed and verified. 
We take responsibility for the final content of this work, including text, claims, or artifacts produced with the aid of generative AI.

\bibliography{AISTATS2027_SC}

\clearpage
\section*{Checklist}



\begin{enumerate}

  \item For all models and algorithms presented, check if you include:
  \begin{enumerate}
    \item A clear description of the mathematical setting, assumptions, algorithm, and/or model. [Yes] 
    \item An analysis of the properties and complexity (time, space, sample size) of any algorithm. [Yes] 
    \item (Optional) Anonymized source code, with specification of all dependencies, including external libraries. [Yes]
  \end{enumerate}

  \item For any theoretical claim, check if you include:
  \begin{enumerate}
    \item Statements of the full set of assumptions of all theoretical results. [Yes]
    \item Complete proofs of all theoretical results. [Yes]
    \item Clear explanations of any assumptions. [Yes]
  \end{enumerate}

  \item For all figures and tables that present empirical results, check if you include:
  \begin{enumerate}
    \item The code, data, and instructions needed to reproduce the main experimental results (either in the supplemental material or as a URL). [Yes]
    \item All the training details (e.g., data splits, hyperparameters, how they were chosen). [Yes]
    \item A clear definition of the specific measure or statistics and error bars (e.g., with respect to the random seed after running experiments multiple times). [Yes]
    \item A description of the computing infrastructure used. (e.g., type of GPUs, internal cluster, or cloud provider). [Yes]
  \end{enumerate}

  \item If you are using existing assets (e.g., code, data, models) or curating/releasing new assets, check if you include:
  \begin{enumerate}
    \item Citations of the creator If your work uses existing assets. [Yes]
    \item The license information of the assets, if applicable. [Yes]
    \item New assets either in the supplemental material or as a URL, if applicable. [Not Applicable]
    \item Information about consent from data providers/curators. [Not Applicable]
    \item Discussion of sensible content if applicable, e.g., personally identifiable information or offensive content. [Not Applicable]
  \end{enumerate}

  \item If you used crowdsourcing or conducted research with human subjects, check if you include:
  \begin{enumerate}
    \item The full text of instructions given to participants and screenshots. [Not Applicable]
    \item Descriptions of potential participant risks, with links to Institutional Review Board (IRB) approvals if applicable. [Not Applicable]
    \item The estimated hourly wage paid to participants and the total amount spent on participant compensation. [Not Applicable]
  \end{enumerate}

\end{enumerate}

\clearpage
\appendix
\thispagestyle{empty}
\onecolumn
\aistatstitle{Supplementary Materials}

\section{Notation} 
\label{Description}

\begin{table*}[h]
    \centering
    \begin{threeparttable}[]
    \begin{tabular}{c|l} 
    \toprule
    Notation  & Description  \\ 	
    \midrule 
    $n$ & the number of vertices  \\ 	
    $m$ & the number of partitions  \\ 	
    $l$ & the number of distinct sublabels   \\ 
    $V$ & $V \coloneqq \{1, \dotsc, n\}$ \\ 
    $W$ & $W \coloneqq \{1, \dotsc, m\}$ \\ 
    $V_\circ$ & $V_\circ \coloneqq \{1, \dotsc, l\}$ \\ 
    $V_j$ & $V_j(\bm{\omega}) \coloneqq \{i \in V : \omega_i = j\}$  \\ 
    $S_k$ & $S_k \coloneqq \{i \in V : \sigma_i = k\}$ for $k \in V_\circ$ \\ 
    $\bm{\omega}$ & $\bm{\omega} \coloneqq (\omega_i)_{i \in V}$, where $\omega_i \in W$ denotes the \emph{label} assigned to vertex $i$ \\   
    $\bm{\sigma}$ & $\bm{\sigma} \coloneqq (\sigma_i)_{i \in V}$, where $\sigma_i \in V_\circ$ denotes the \emph{sublabel} assigned to vertex $i$ \\ 
    $f$ & $f : \mathcal{X} \to \mathbb{R}$ \\ 
    $g$ & $g(\bm{\omega}; \bm{y}) \coloneqq (|V_{\omega_i}(\bm{\omega})|^{-1} \sum_{i^\prime \in V_{\omega_i}(\bm{\omega})} y_{i^\prime})_{i \in V}$  \\ 
    $\bm{f}$ & $\bm{f} \coloneqq (f(\bm{x}_i))_{i \in V}$ with $\bm{x}_i \in \mathcal{X}$  \\   
    $\bm{f}_\sigma$ & $\bm{f}_\sigma \coloneqq g(\bm{\sigma}; \bm{f})$ \\ 
    $\bm{f}_p$ & $\bm{f}_p(S_\circ) \coloneqq ([\bm{f}]_i \mathbf{1}_{\{\sigma_i \notin S_\circ\}} + [\bm{f}_\sigma]_i \mathbf{1}_{\{\sigma_i \in S_\circ\}})_{i \in V}$ for $S_\circ \subseteq V_\circ$ \\   
    $\eta$ & $\eta(S_\circ) \coloneqq \|\bm{f}_p(S_\circ) - g(\underline{\bm{\omega}}; \bm{f})\|_2 - \|\bm{f} - \bm{f}_p(S_\circ)\|_2$ for $S_\circ \subseteq V_\circ$ \\ 
    $\epsilon$ & $\epsilon(\underline{\bm{\omega}}, \overline{\bm{\omega}}) \coloneqq \|\bm{f} - g(\underline{\bm{\omega}}; \bm{f})\|_2 - \|\bm{f} - g(\overline{\bm{\omega}}; \bm{f})\|_2$ \\   
    $\delta$ & $\delta \coloneqq 2 \|\bm{f} - \bm{f}_\sigma\|_2$ \\ 
    $\zeta$ & $\zeta \coloneqq \|\bm{f} - g(\underline{\bm{\omega}}; \bm{f})\|_2 - \min_{S_\circ \in \mathcal{B}_\circ} \eta(S_\circ)$ with $\mathcal{B}_\circ \coloneqq \{B_\circ(\bm{\omega}) : \bm{\omega} \in \Omega\}$ \\ 
    $\Omega_0$ & $\Omega_0 \coloneqq \{\bm{\omega} \in W^n : G[V_j(\bm{\omega})] \text{ is empty or connected for each } j \in W\}$ \\ 
    $\Omega$ & See \Cref{def_omega} \\ 
    $\Omega^\prime$ & See \Cref{lem_omega_def} \\ 
    $\Omega^{\prime\prime}$ & See \Cref{feasible_set_def} \\ 
    $\Omega_+$ & See \Cref{Omega_plus} \\ 
    $\overline{\Omega}$ & See \Cref{def_omega1} \\ 
    $\underline{\Omega}$ & $\underline{\Omega} \coloneqq \{\bm{\omega} \in \Omega_0 : |\{\omega_i : i \in S_k\}| = 1 \text{ for each } k \in V_\circ\}$  \\ 
    $\widetilde{\Omega}$ & $\widetilde{\Omega} \coloneqq \{\bm{\omega} \in \Omega : C_j^\circ(\bm{\omega}) \subseteq C_j^\circ(\underline{\bm{\omega}}) \text{ for each } j \in W\}$  \\ 
    $\widetilde{\Omega}_+$ & $\widetilde{\Omega}_+ \coloneqq \{\bm{\omega} \in \Omega_+ : C_j^\circ(\bm{\omega}) \subseteq C_j^\circ(\underline{\bm{\omega}}) \text{ for each } j \in W\}$  \\ 
    $\widetilde{\mathcal{B}}_\circ$ & $\widetilde{\mathcal{B}}_\circ \coloneqq \{B_\circ(\bm{\omega}) : \bm{\omega} \in \widetilde{\Omega}\}$ \\ 
    $\widetilde{\mathcal{B}}_\circ^+$ & $\widetilde{\mathcal{B}}_\circ^+ \coloneqq \{B_\circ(\bm{\omega}) : \bm{\omega} \in \widetilde{\Omega}_+\}$ \\ 
    $\Omega_p$ & $\Omega_p(\bm{\omega}) \coloneqq \{\bm{\omega}_p \in \Omega : C_j^\circ(\bm{\omega}) \subseteq C_j^\circ(\bm{\omega}_p) \text{ for each } j \in W\}$  \\ 
    $\overline{\bm{\omega}}$ & an optimal label assignment vector for $\mathrm{GPP}(\overline{\Omega}, \bm{f})$ \\   
    $\underline{\bm{\omega}}$ & an optimal label assignment vector for $\mathrm{GPP}(\underline{\Omega}, \bm{f}_\sigma)$ \\   
    $\bm{\omega}_*$ & an optimal label assignment vector for $\mathrm{GPP}(\Omega, \bm{f})$ \\   
    $C_\circ$ & $C_\circ(\bm{\omega}) \coloneqq \bigcup_{j \in W} C_j^\circ(\bm{\omega})$  \\ 
    $C_j^\circ$ & $C_j^\circ(\bm{\omega}) \coloneqq \{\sigma_i : i \in C_j(\bm{\omega})\}$  \\  
    $C_j$ & $C_j(\bm{\omega}) \coloneqq \bigcup_{k \in V_\circ : S_k \subseteq V_j(\bm{\omega})} S_k$  \\ 
    $B_\circ$ & $B_\circ(\bm{\omega}) \coloneqq \bigcup_{j \in W} B_j^\circ(\bm{\omega})$  \\  
    $B_j^\circ$ & $B_j^\circ(\bm{\omega}) \coloneqq \{\sigma_i : i \in V_j(\bm{\omega}) \setminus C_j(\bm{\omega})\}$  \\  
    $\mathcal{B}_j$ & $\mathcal{B}_j(\bm{\omega}) \coloneqq \{S_k \cap V_j(\bm{\omega}) : k \in V_\circ \text{ satisfying } S_k \not\subseteq V_j(\bm{\omega})\} \setminus \{\emptyset\}$ \\  
    \bottomrule 
    \end{tabular}
    \end{threeparttable}
\end{table*}

\newpage
\section{Proofs} 
\label{Proofs}

\subsection{Proof of \Cref{aggregation}} 
By the definition of $\bm{f}_\sigma$, for each $\bm{\omega} \in \underline{\Omega}$, we obtain  
\begin{align}  
(\bm{f} - \bm{f}_\sigma)^\top (\bm{f}_\sigma - g(\bm{\omega}; \bm{f}_\sigma)) = \bm{0}. 
\end{align} 
Therefore, for each $\bm{\omega} \in \underline{\Omega}$, we have 
\begin{align} 
\|\bm{f} - g(\bm{\omega}; \bm{f})\|_2^2 
&= \|\bm{f} - g(\bm{\omega}; \bm{f}_\sigma)\|_2^2& \nonumber \\ 
&= \|\bm{f} - \bm{f}_\sigma + \bm{f}_\sigma - g(\bm{\omega}; \bm{f}_\sigma)\|_2^2& \nonumber \\ 
&= \|\bm{f} - \bm{f}_\sigma\|_2^2 + 2 (\bm{f} - \bm{f}_\sigma)^\top (\bm{f}_\sigma - g(\bm{\omega}; \bm{f}_\sigma)) + \|\bm{f}_\sigma - g(\bm{\omega}; \bm{f}_\sigma)\|_2^2& \nonumber \\  
&= \|\bm{f} - \bm{f}_\sigma\|_2^2 + \|\bm{f}_\sigma - g(\bm{\omega}; \bm{f}_\sigma)\|_2^2.&  
\end{align} 
Consequently, the set of optimal label assignment vectors for $\mathrm{GPP}(\underline{\Omega}, \bm{f})$ coincides with that for $\mathrm{GPP}(\underline{\Omega}, \bm{f}_\sigma)$.

\subsection{Proof of \Cref{trivial_relation}} 
For each $\bm{\omega} \in \underline{\Omega}$, 
\begin{align} 
|\{\omega_i : i \in V, \sigma_i = 1\}| = \cdots = |\{\omega_i : i \in V, \sigma_i = l\}| = 1. 
\end{align} 
Thus, the label reassignment operation in \Cref{def_omega1} leaves $\bm{\omega}$ unchanged. 
Consequently, we have $\underline{\Omega} \subseteq \overline{\Omega}$.

\subsection{Proof of \Cref{identical_solution}} 
\label{identical_solution_proof} 
Since $\bm{f}_\sigma$ is piecewise constant over the sublabel groups, we consider the following reassignment. 
First, keep the cluster means induced by the original assignment $\bm{\omega} \in \overline{\Omega}$ fixed. 
For each sublabel group, assign all of its vertices to one of the labels already appearing in that group that minimizes the residual with respect to these fixed cluster means. 
Since $\bm{f}_\sigma$ is constant within each sublabel group, this reassignment does not increase the objective. 
After performing this reassignment for all sublabel groups, we obtain $\bm{\omega}^\prime \in \underline{\Omega}$. 
We then recompute the cluster means for the resulting partition. 
Since each sublabel group is assigned a common label under $\bm{\omega}^\prime$, we have $g(\bm{\omega}^\prime; \bm{f}) = g(\bm{\omega}^\prime; \bm{f}_\sigma)$. 
Hence, recomputing the cluster means cannot increase the objective. 
Therefore,
\begin{align} 
\|\bm{f}_\sigma - g(\bm{\omega}; \bm{f})\|_2 
\geq \|\bm{f}_\sigma - (|V_{[\bm{\omega}^\prime]_{i^\prime}}(\bm{\omega})|^{-1} \sum_{i \in V_{[\bm{\omega}^\prime]_{i^\prime}}(\bm{\omega})} f(\bm{x}_i))_{i^\prime \in V}\|_2 
\geq \|\bm{f}_\sigma - g(\bm{\omega}^\prime; \bm{f})\|_2.   
\end{align} 

Suppose, to the contrary, that \Cref{GPP_semi} admits no optimal label assignment vector in $\underline{\Omega}$. 
By the above argument, there exists a reassignment vector $\bm{\omega}^\prime \in \underline{\Omega}$ whose objective value is no larger than that of an optimal label assignment vector $\bm{\omega}$ for \Cref{GPP_semi}. 
Hence, $\bm{\omega}^\prime$ is also optimal for \Cref{GPP_semi}, which contradicts the supposition. 
Therefore, \Cref{GPP_semi} admits an optimal label assignment vector in $\underline{\Omega}$. 
Accordingly, we have 
\begin{align} 
\min_{\bm{\omega} \in \overline{\Omega}} \|\bm{f}_\sigma - g(\bm{\omega}; \bm{f})\|_2 
= \min_{\bm{\omega} \in \underline{\Omega}} \|\bm{f}_\sigma - g(\bm{\omega}; \bm{f})\|_2  
= \min_{\bm{\omega} \in \underline{\Omega}} \|\bm{f}_\sigma - g(\bm{\omega}; \bm{f}_\sigma)\|_2.  
\end{align} 
Consequently, each optimal label assignment vector for $\mathrm{GPP}(\underline{\Omega}, \bm{f}_\sigma)$ is also optimal for \Cref{GPP_semi}.

\subsection{Proof of \Cref{guarantee_underline}} 
By the definition of $\underline{\bm{\omega}}$ and $\overline{\bm{\omega}}$, we have 
\begin{align} 
\|\bm{f} - g(\overline{\bm{\omega}}; \bm{f})\|_2 \leq \|\bm{f} - g(\underline{\bm{\omega}}; \bm{f})\|_2. 
\end{align} 
By \Cref{identical_solution}, we obtain  
\begin{align} 
\|\bm{f}_\sigma - g(\underline{\bm{\omega}}; \bm{f})\|_2 \leq \|\bm{f}_\sigma - g(\overline{\bm{\omega}}; \bm{f})\|_2. 
\end{align} 
Accordingly, we have  
\begin{align} 
\|\bm{f} - g(\overline{\bm{\omega}}; \bm{f})\|_2 
&\geq \|(- \bm{f}_\sigma + g(\overline{\bm{\omega}}; \bm{f})) - (\bm{f} - \bm{f}_\sigma)\|_2& \nonumber \\  
&\geq \|\bm{f}_\sigma - g(\overline{\bm{\omega}}; \bm{f})\|_2 - \|\bm{f} - \bm{f}_\sigma\|_2& \nonumber \\  
&\geq \|\bm{f}_\sigma - g(\underline{\bm{\omega}}; \bm{f})\|_2 - \|\bm{f} - \bm{f}_\sigma\|_2& \nonumber \\  
&\geq \|(- \bm{f} + g(\underline{\bm{\omega}}; \bm{f})) - (\bm{f}_\sigma - \bm{f})\|_2 - \|\bm{f} - \bm{f}_\sigma\|_2& \nonumber \\  
&\geq \|\bm{f} - g(\underline{\bm{\omega}}; \bm{f})\|_2 - \|\bm{f} - \bm{f}_\sigma\|_2 - \|\bm{f} - \bm{f}_\sigma\|_2& \nonumber \\  
&= \|\bm{f} - g(\underline{\bm{\omega}}; \bm{f})\|_2 - 2 \|\bm{f} - \bm{f}_\sigma\|_2& \nonumber \\   
&= \|\bm{f} - g(\underline{\bm{\omega}}; \bm{f})\|_2 - \delta.& 
\end{align} 
Consequently, the following holds: 
\begin{align} 
0 \leq \epsilon(\underline{\bm{\omega}}, \overline{\bm{\omega}}) \leq \delta. 
\end{align}

\subsection{Proof of \Cref{lem_omega}}  
In the characterization of \Cref{lem_omega_def}, for each $k \in V_\circ$, the vertices in the $k$-th sublabel group satisfying $|\{\omega_i : i \in S_k\}| > 1$ can be reassigned to some label different from their current labels. 
Therefore, the set $\Omega^\prime$ is the largest subset of $\Omega_0$ such that, 
for each $\bm{\omega} \in \Omega^\prime$ and each $j \in W$, 
$G[V_j(\bm{\omega}) \setminus \bigcup_{k \in S_\circ} S_k]$ is empty or connected for every $S_\circ \subseteq V_\circ$ satisfying $|\{\omega_i : i \in S_k\}| > 1$ for all $k \in S_\circ$. 
Moreover, \Cref{def_omega} is unchanged if replacing $\mathcal{S} \subseteq \mathcal{B}_j(\bm{\omega})$ with 
\begin{align} 
\mathcal{S} 
\in 2^{\{S_k : k \in V_\circ \text{ satisfying } |\{\omega_i : i \in S_k\}| > 1\}} 
= \{\{S_k : k \in S_\circ\} : S_\circ \subseteq V_\circ \text{ satisfying } |\{\omega_i : i \in S_k\}| > 1 \text{ for all } k \in S_\circ\}. 
\end{align} 
Consequently, we have $\Omega = \Omega^\prime$.

\subsection{Proof of \Cref{lem_omega21}}
For each $\bm{\omega} \in \underline{\Omega} \subseteq \Omega_0$ and each $j \in W$, 
\begin{align} 
G[V_j(\bm{\omega}) \setminus \bigcup_{S \in \mathcal{S}} S] = G[V_j(\bm{\omega})] 
\end{align} 
is empty or connected for every $\mathcal{S} \subseteq \mathcal{B}_j(\bm{\omega}) = \emptyset$. 
Therefore, we have $\underline{\Omega} \subseteq \Omega$. 

The difference between $\overline{\Omega}$ and $\Omega^\prime$ lies in whether the reassignment condition is required only for $S_\circ = V_\circ$ or for every $S_\circ \subseteq V_\circ$. 
Therefore, by \Cref{lem_omega}, we have $\Omega = \Omega^\prime \subseteq \overline{\Omega}$. 
Consequently, we have $\underline{\Omega} \subseteq \Omega \subseteq \overline{\Omega}$.

\subsection{Proof of \Cref{guarantee0}}
By \Cref{lem_omega21}, we have 
\begin{align} 
\|\bm{f} - g(\overline{\bm{\omega}}; \bm{f})\|_2 \leq \|\bm{f} - g(\bm{\omega}_*; \bm{f})\|_2 \leq \|\bm{f} - g(\underline{\bm{\omega}}; \bm{f})\|_2. 
\end{align} 
Moreover, by \Cref{guarantee_underline}, we obtain 
\begin{align} 
\|\bm{f} - g(\underline{\bm{\omega}}; \bm{f})\|_2 - \|\bm{f} - g(\bm{\omega}_*; \bm{f})\|_2 \leq \|\bm{f} - g(\underline{\bm{\omega}}; \bm{f})\|_2 - \|\bm{f} - g(\overline{\bm{\omega}}; \bm{f})\|_2 \leq \delta. 
\end{align} 
Consequently, the following holds:   
\begin{align} 
0 \leq \epsilon(\underline{\bm{\omega}}, \bm{\omega}_*) \leq \delta. 
\end{align}

\subsection{Proof of \Cref{identical_solution_p}}  
The proof follows the same reassignment argument as \Cref{identical_solution_proof}. 
By the definition of $\Omega_p$, for each $\bm{\omega}_p \in \Omega_p(\bm{\omega})$, 
applying the reassignment operation to all sublabel groups yields a reassignment vector $\bm{\omega}^\prime \in \underline{\Omega} \cap \Omega_p(\bm{\omega})$ such that 
\begin{align} 
\|\bm{f}_\sigma - g(\bm{\omega}_p; \bm{f})\|_2 
\geq \|\bm{f}_\sigma - (|V_{[\bm{\omega}^\prime]_{i^\prime}}(\bm{\omega}_p)|^{-1} \sum_{i \in V_{[\bm{\omega}^\prime]_{i^\prime}}(\bm{\omega}_p)} f(\bm{x}_i))_{i^\prime \in V}\|_2 
\geq \|\bm{f}_\sigma - g(\bm{\omega}^\prime; \bm{f})\|_2.   
\end{align} 
Therefore, \Cref{GPP_semi_p} admits an optimal label assignment vector in $\underline{\Omega} \cap \Omega_p(\bm{\omega})$. 
Hence, we have 
\begin{align} 
\min_{\bm{\omega}_p \in \Omega_p(\bm{\omega})} \|\bm{f}_\sigma - g(\bm{\omega}_p; \bm{f})\|_2 
= \min_{\bm{\omega}_p \in \underline{\Omega} \cap \Omega_p(\bm{\omega})} \|\bm{f}_\sigma - g(\bm{\omega}_p; \bm{f})\|_2 
= \min_{\bm{\omega}_p \in \underline{\Omega} \cap \Omega_p(\bm{\omega})} \|\bm{f}_\sigma - g(\bm{\omega}_p; \bm{f}_\sigma)\|_2.   
\end{align} 
Consequently, each optimal label assignment vector for $\mathrm{GPP}(\underline{\Omega} \cap \Omega_p(\bm{\omega}), \bm{f}_\sigma)$ is also optimal for \Cref{GPP_semi_p}.

\subsection{Proof of \Cref{LB_sub}}  
By \Cref{identical_solution_p}, for each $\bm{\omega} \in \Omega$, we obtain 
\begin{align} 
\|\bm{f}_\sigma - g(\bm{\omega}_p^*(\bm{\omega}); \bm{f})\|_2 \leq \|\bm{f}_\sigma - g(\bm{\omega}_*^p(\bm{\omega}); \bm{f})\|_2, 
\end{align} 
where $\bm{\omega}_*^p(\bm{\omega})$ and $\bm{\omega}_p^*(\bm{\omega})$ are optimal label assignment vectors of $\mathrm{GPP}(\Omega_p(\bm{\omega}), \bm{f})$ and $\mathrm{GPP}(\underline{\Omega} \cap \Omega_p(\bm{\omega}), \bm{f}_\sigma)$, respectively.   
By the definition of $\bm{f}_\sigma$, $\bm{f}_p$, $\bm{\omega}_*^p$, and $\underline{\bm{\omega}}$, for each $\bm{\omega} \in \Omega$, we have 
\begin{align}  
(\bm{f}_p(B_\circ(\bm{\omega})) - \bm{f}_\sigma)^\top (\bm{f}_\sigma - g(\bm{\omega}_*^p(\bm{\omega}); \bm{f})) = \bm{0}, \\ 
(\bm{f}_p(B_\circ(\bm{\omega})) - \bm{f}_\sigma)^\top (\bm{f}_\sigma - g(\underline{\bm{\omega}}; \bm{f})) = \bm{0}. 
\end{align} 
Therefore, for each $\bm{\omega} \in \Omega$, the following holds: 
\begin{align} 
&\|\bm{f} - g(\bm{\omega}_*^p(\bm{\omega}); \bm{f})\|_2& \nonumber \\  
&= \|(- \bm{f}_p(B_\circ(\bm{\omega})) + g(\bm{\omega}_*^p(\bm{\omega}); \bm{f})) - (\bm{f} - \bm{f}_p(B_\circ(\bm{\omega})))\|_2& \nonumber \\  
&\geq \|\bm{f}_p(B_\circ(\bm{\omega})) - g(\bm{\omega}_*^p(\bm{\omega}); \bm{f})\|_2 - \|\bm{f} - \bm{f}_p(B_\circ(\bm{\omega}))\|_2& \nonumber \\ 
&= \|\bm{f}_p(B_\circ(\bm{\omega})) - \bm{f}_\sigma + \bm{f}_\sigma - g(\bm{\omega}_*^p(\bm{\omega}); \bm{f})\|_2 - \|\bm{f} - \bm{f}_p(B_\circ(\bm{\omega}))\|_2& \nonumber \\ 
&= \sqrt{\|\bm{f}_p(B_\circ(\bm{\omega})) - \bm{f}_\sigma\|_2^2 + 2 (\bm{f}_p(B_\circ(\bm{\omega})) - \bm{f}_\sigma)^\top (\bm{f}_\sigma - g(\bm{\omega}_*^p(\bm{\omega}); \bm{f})) + \|\bm{f}_\sigma - g(\bm{\omega}_*^p(\bm{\omega}); \bm{f})\|_2^2} - \|\bm{f} - \bm{f}_p(B_\circ(\bm{\omega}))\|_2& \nonumber \\ 
&= \sqrt{\|\bm{f}_p(B_\circ(\bm{\omega})) - \bm{f}_\sigma\|_2^2 + \|\bm{f}_\sigma - g(\bm{\omega}_*^p(\bm{\omega}); \bm{f})\|_2^2} - \|\bm{f} - \bm{f}_p(B_\circ(\bm{\omega}))\|_2& \nonumber \\ 
&\geq \sqrt{\|\bm{f}_p(B_\circ(\bm{\omega})) - \bm{f}_\sigma\|_2^2 + \|\bm{f}_\sigma - g(\bm{\omega}_p^*(\bm{\omega}); \bm{f})\|_2^2} - \|\bm{f} - \bm{f}_p(B_\circ(\bm{\omega}))\|_2& \nonumber \\ 
&= \sqrt{\|\bm{f}_p(B_\circ(\bm{\omega})) - \bm{f}_\sigma\|_2^2 + \|\bm{f}_\sigma - g(\bm{\omega}_p^*(\bm{\omega}); \bm{f}_\sigma)\|_2^2} - \|\bm{f} - \bm{f}_p(B_\circ(\bm{\omega}))\|_2& \nonumber \\   
&\geq \sqrt{\|\bm{f}_p(B_\circ(\bm{\omega})) - \bm{f}_\sigma\|_2^2 + \|\bm{f}_\sigma - g(\underline{\bm{\omega}}; \bm{f})\|_2^2} - \|\bm{f} - \bm{f}_p(B_\circ(\bm{\omega}))\|_2& \nonumber \\ 
&= \sqrt{\|\bm{f}_p(B_\circ(\bm{\omega})) - \bm{f}_\sigma\|_2^2 + 2 (\bm{f}_p(B_\circ(\bm{\omega})) - \bm{f}_\sigma)^\top (\bm{f}_\sigma - g(\underline{\bm{\omega}}; \bm{f})) + \|\bm{f}_\sigma - g(\underline{\bm{\omega}}; \bm{f})\|_2^2} - \|\bm{f} - \bm{f}_p(B_\circ(\bm{\omega}))\|_2& \nonumber \\ 
&= \|\bm{f}_p(B_\circ(\bm{\omega})) - \bm{f}_\sigma + \bm{f}_\sigma - g(\underline{\bm{\omega}}; \bm{f})\|_2 - \|\bm{f} - \bm{f}_p(B_\circ(\bm{\omega}))\|_2& \nonumber \\ 
&= \|\bm{f}_p(B_\circ(\bm{\omega})) - g(\underline{\bm{\omega}}; \bm{f})\|_2 - \|\bm{f} - \bm{f}_p(B_\circ(\bm{\omega}))\|_2& \nonumber \\ 
&= \eta(B_\circ(\bm{\omega})).& 
\end{align} 
Consequently, for each $\bm{\omega} \in \Omega$, we have 
\begin{align} 
\epsilon(\underline{\bm{\omega}}, \bm{\omega})
&= \|\bm{f} - g(\underline{\bm{\omega}}; \bm{f})\|_2 - \|\bm{f} - g(\bm{\omega}; \bm{f})\|_2 \nonumber \\ 
&\leq \|\bm{f} - g(\underline{\bm{\omega}}; \bm{f})\|_2 - \min_{\bm{\omega}^\prime \in \Omega : \mathcal{B}_\circ(\bm{\omega}^\prime) = \mathcal{B}_\circ(\bm{\omega})} \|\bm{f} - g(\bm{\omega}_*^p(\bm{\omega}^\prime); \bm{f})\|_2 \nonumber \\ 
&\leq \|\bm{f} - g(\underline{\bm{\omega}}; \bm{f})\|_2 - \min_{\bm{\omega}^\prime \in \Omega : \mathcal{B}_\circ(\bm{\omega}^\prime) = \mathcal{B}_\circ(\bm{\omega})} \eta(B_\circ(\bm{\omega}^\prime)) \nonumber \\ 
&= \|\bm{f} - g(\underline{\bm{\omega}}; \bm{f})\|_2 - \eta(B_\circ(\bm{\omega})).  
\end{align}

\subsection{Proof of \Cref{LB}}  
By \Cref{LB_sub}, we have 
\begin{align} 
\epsilon(\underline{\bm{\omega}}, \bm{\omega}_*)
&= \|\bm{f} - g(\underline{\bm{\omega}}; \bm{f})\|_2 - \|\bm{f} - g(\bm{\omega}_*; \bm{f})\|_2 \nonumber \\ 
&= \|\bm{f} - g(\underline{\bm{\omega}}; \bm{f})\|_2 - \min_{\bm{\omega} \in \Omega} \|\bm{f} - g(\bm{\omega}; \bm{f})\|_2 \nonumber \\ 
&\leq \|\bm{f} - g(\underline{\bm{\omega}}; \bm{f})\|_2 - \min_{\bm{\omega} \in \Omega} \eta(B_\circ(\bm{\omega})) \nonumber \\ 
&= \|\bm{f} - g(\underline{\bm{\omega}}; \bm{f})\|_2 - \min_{S_\circ \in \mathcal{B}_\circ} \eta(S_\circ) \nonumber \\ 
&= \zeta. 
\end{align}

\subsection{Proof of \Cref{monotonicity}}  
Consider a set $S_\circ \subseteq V_\circ$. 
By the definition of $\bm{f}_p$ and $\underline{\bm{\omega}}$, we have 
\begin{align} 
(\bm{f}_p(S_\circ) - \bm{f})^\top (\bm{f}_p(S_\circ) - g(\underline{\bm{\omega}}; \bm{f})) = \bm{0}. 
\end{align} 
Therefore, for each $\bm{\omega} \in \Omega$, the following holds: 
\begin{align} 
\eta(S_\circ) 
&= \|\bm{f}_p(S_\circ) - g(\underline{\bm{\omega}}; \bm{f})\|_2 - \|\bm{f} - \bm{f}_p(S_\circ)\|_2 \nonumber \\ 
&= \sqrt{\|\bm{f}_p(S_\circ) - g(\underline{\bm{\omega}}; \bm{f})\|_2^2} - \|\bm{f} - \bm{f}_p(S_\circ)\|_2 \nonumber \\ 
&= \sqrt{\|\bm{f}_p(S_\circ) - \bm{f} + \bm{f} - g(\underline{\bm{\omega}}; \bm{f})\|_2^2} - \|\bm{f} - \bm{f}_p(S_\circ)\|_2 \nonumber \\ 
&= \sqrt{\|\bm{f}_p(S_\circ) - \bm{f}\|_2^2 + 2 (\bm{f}_p(S_\circ) - \bm{f})^\top (\bm{f} - g(\underline{\bm{\omega}}; \bm{f})) + \|\bm{f} - g(\underline{\bm{\omega}}; \bm{f})\|_2^2} - \|\bm{f} - \bm{f}_p(S_\circ)\|_2 \nonumber \\ 
&= \sqrt{\|\bm{f}_p(S_\circ) - \bm{f}\|_2^2 + 2 (\bm{f}_p(S_\circ) - \bm{f})^\top (\bm{f} - \bm{f}_p(S_\circ) + \bm{f}_p(S_\circ) - g(\underline{\bm{\omega}}; \bm{f})) + \|\bm{f} - g(\underline{\bm{\omega}}; \bm{f})\|_2^2} - \|\bm{f} - \bm{f}_p(S_\circ)\|_2 \nonumber \\ 
&= \sqrt{2 (\bm{f}_p(S_\circ) - \bm{f})^\top (\bm{f}_p(S_\circ) - g(\underline{\bm{\omega}}; \bm{f})) + \|\bm{f} - g(\underline{\bm{\omega}}; \bm{f})\|_2^2 - \|\bm{f}_p(S_\circ) - \bm{f}\|_2^2} - \|\bm{f} - \bm{f}_p(S_\circ)\|_2 \nonumber \\ 
&= \sqrt{\|\bm{f} - g(\underline{\bm{\omega}}; \bm{f})\|_2^2 - \|\bm{f}_p(S_\circ) - \bm{f}\|_2^2} - \|\bm{f} - \bm{f}_p(S_\circ)\|_2 \nonumber \\  
&= \sqrt{\|\bm{f} - g(\underline{\bm{\omega}}; \bm{f})\|_2^2 - \|\bm{f} - \bm{f}_p(S_\circ)\|_2^2} - \|\bm{f} - \bm{f}_p(S_\circ)\|_2. 
\end{align} 
Consequently, for $S_\circ \subseteq V_\circ$, $\eta(S_\circ)$ is strictly decreasing in $\|\bm{f} - \bm{f}_p(S_\circ)\|_2$.

\subsection{Proof of \Cref{tight_result}}  
The following holds: 
\begin{align} 
\zeta 
&= \|\bm{f} - g(\underline{\bm{\omega}}; \bm{f})\|_2 - \min_{S_\circ \in \mathcal{B}_\circ} \eta(S_\circ) \nonumber \\ 
&= \|\bm{f} - g(\underline{\bm{\omega}}; \bm{f})\|_2 - \min_{S_\circ \in \mathcal{B}_\circ} \bigl(\|\bm{f}_p(S_\circ) - g(\underline{\bm{\omega}}; \bm{f})\|_2 - \|\bm{f} - \bm{f}_p(S_\circ)\|_2\bigr) \nonumber \\  
&= \|\bm{f} - g(\underline{\bm{\omega}}; \bm{f})\|_2 - \min_{S_\circ \in \mathcal{B}_\circ} \bigl(\|- (- \bm{f}_p(S_\circ) + \bm{f}) + (\bm{f} - g(\underline{\bm{\omega}}; \bm{f}))\|_2 - \|\bm{f} - \bm{f}_p(S_\circ)\|_2\bigr) \nonumber \\  
&\leq \|\bm{f} - g(\underline{\bm{\omega}}; \bm{f})\|_2 - \min_{S_\circ \in \mathcal{B}_\circ} \bigl(- \|- \bm{f}_p(S_\circ) + \bm{f}\|_2 + \|\bm{f} - g(\underline{\bm{\omega}}; \bm{f})\|_2 - \|\bm{f} - \bm{f}_p(S_\circ)\|_2\bigr) \nonumber \\  
&= \|\bm{f} - g(\underline{\bm{\omega}}; \bm{f})\|_2 - \min_{S_\circ \in \mathcal{B}_\circ} \bigl(\|\bm{f} - g(\underline{\bm{\omega}}; \bm{f})\|_2 - 2 \|\bm{f} - \bm{f}_p(S_\circ)\|_2\bigr) \nonumber \\  
&= 2 \max_{S_\circ \in \mathcal{B}_\circ} \|\bm{f} - \bm{f}_p(S_\circ)\|_2 \nonumber \\ 
&\leq \delta.   
\end{align}

\subsection{Proof of \Cref{feasible_set}}  
For each $\bm{\omega} \in \Omega$ and each $j \in W$,  
\begin{align} 
G[C_j(\bm{\omega})] = G[V_j(\bm{\omega}) \setminus \bigcup_{S \in \mathcal{B}_j(\bm{\omega})} S] 
\end{align} 
is empty or connected, and if $C_j(\bm{\omega}) = \emptyset$, then 
\begin{align} 
G[\bigcup_{S \in \mathcal{S}} S] = G[C_j(\bm{\omega}) \cup \bigcup_{S \in \mathcal{S}} S] = G[V_j(\bm{\omega}) \setminus \bigcup_{S \in \mathcal{B}_j(\bm{\omega}) \setminus \mathcal{S}} S]   
\end{align} 
is empty or connected for every $\mathcal{S} \subseteq \mathcal{B}_j(\bm{\omega})$; otherwise,  
\begin{align} 
G[S \cup C_j(\bm{\omega})] = G\bigl[V_j(\bm{\omega}) \setminus \bigcup_{S^\prime \in \mathcal{B}_j(\bm{\omega}) \setminus S} S^\prime\bigr] 
\end{align} 
is connected for all $S \in \mathcal{B}_j(\bm{\omega})$. 
Hence, we have $\Omega \subseteq \Omega^{\prime\prime}$. 
Moreover, for each $\bm{\omega} \in \Omega^{\prime\prime}$ and each $j \in W$,  
\begin{align} 
G[V_j(\bm{\omega}) \setminus \bigcup_{S \in \mathcal{S}} S] 
= G[(C_j(\bm{\omega}) \cup \bigcup_{S \in \mathcal{B}_j(\bm{\omega})} S) \setminus \bigcup_{S \in \mathcal{S}} S] 
= G[C_j(\bm{\omega}) \cup \bigcup_{S \in \mathcal{B}_j(\bm{\omega}) \setminus \mathcal{S}} S] 
\end{align} 
is empty or connected for every $\mathcal{S} \subseteq \mathcal{B}_j(\bm{\omega})$.  
Therefore, we have $\Omega^{\prime\prime} \subseteq \Omega$. 
As a result, we obtain $\Omega = \Omega^{\prime\prime}$.

\subsection{Proof of \Cref{feasible_set_core}}  
For each $\bm{\omega} \in \Omega$ and each $j \in W$,  
\begin{align} 
G_\circ[C_j^\circ(\bm{\omega})] 
= \bigl(C_j^\circ(\bm{\omega}), E_\circ\bigl(C_j(\bm{\omega})\bigr)\bigr)  
= \bigl(C_j^\circ(\bm{\omega}), E_\circ\bigl(V_j(\bm{\omega}) \setminus \bigcup_{S \in \mathcal{B}_j(\bm{\omega})} S\bigr)\bigr) 
\end{align} 
is empty or connected, and if $C_j^\circ(\bm{\omega}) = \emptyset$, then 
\begin{align} 
G_\circ[B_j^\circ(\bm{\omega})] 
= \bigl(B_j^\circ(\bm{\omega}), E_\circ(\bigcup_{S \in \mathcal{B}_j(\bm{\omega})} S)\bigr)   
= \bigl(B_j^\circ(\bm{\omega}), \bigcup_{\mathcal{S} \subseteq \mathcal{B}_j(\bm{\omega}) : |\mathcal{S}| = |\mathcal{B}_j(\bm{\omega})| - 2} E_\circ\bigl(V_j(\bm{\omega}) \setminus \bigcup_{S \in \mathcal{S}} S\bigr)\bigr)  
\end{align} 
is empty or complete; otherwise, 
\begin{align} 
G_\circ[\{k\} \cup C_j^\circ(\bm{\omega})] 
= \bigl(\{k\} \cup C_j^\circ(\bm{\omega}), E_\circ\bigl(S_k \cup C_j(\bm{\omega})\bigr)\bigr) 
= \bigl(\{k\} \cup C_j^\circ(\bm{\omega}), E_\circ\bigl(V_j(\bm{\omega}) \setminus \bigcup_{S \in \mathcal{B}_j(\bm{\omega}) \setminus \{S_k\}} S\bigr)\bigr) 
\end{align} 
is connected for each $k \in B_j^\circ(\bm{\omega})$. 
Therefore, by \Cref{feasible_set}, we have $\Omega \subseteq \Omega_+$.  
Consequently, it follows that 
\begin{align} 
\widetilde{\mathcal{B}}_\circ 
&= \{B_\circ(\bm{\omega}) : \bm{\omega} \in \widetilde{\Omega}\} \nonumber \\  
&= \{B_\circ(\bm{\omega}) : \bm{\omega} \in \Omega \text{ satisfying } C_j^\circ(\bm{\omega}) \subseteq C_j^\circ(\underline{\bm{\omega}}) \text{ for each } j \in W\} \nonumber \\  
&\subseteq \{B_\circ(\bm{\omega}) : \bm{\omega} \in \Omega_+ \text{ satisfying } C_j^\circ(\bm{\omega}) \subseteq C_j^\circ(\underline{\bm{\omega}}) \text{ for each } j \in W\} \nonumber \\  
&= \{B_\circ(\bm{\omega}) : \bm{\omega} \in \widetilde{\Omega}_+\} \nonumber \\  
&= \widetilde{\mathcal{B}}_\circ^+  
\end{align}

\subsection{Proof of \Cref{BB_equal}} 
Consider a set $U$ in \Cref{splittability}. 
For each $\bm{\omega} \in \widetilde{\Omega}_+$, there exists $\bm{\omega}^\prime \in \widetilde{\Omega}_+$ such that 
\begin{align} 
C_j^\circ(\bm{\omega}) = C_j^\circ(\bm{\omega}^\prime) \text{ for each } j \in W, \\ 
\{[\bm{\omega}^\prime]_i : i \in V, \sigma_i = b\} \subseteq \{\omega_i : i \in V, \sigma_i = b\} \text{ for each } b \in B_\circ(\bm{\omega}), \\ 
|\{[\bm{\omega}^\prime]_i : i \in U, \sigma_i = b\}| = 1 \text{ for each } b \in B_\circ(\bm{\omega}), \\   
|\{[\bm{\omega}^\prime]_i : i \in V \setminus U, \sigma_i = b\}| = 1 \text{ for each } b \in B_\circ(\bm{\omega}).  
\end{align} 
By the definition of $\widetilde{\Omega}_+$, for each $j \in W$,  
\begin{itemize}
\item $G_\circ[C_j^\circ(\bm{\omega}^\prime)]$ is empty or connected,   
\item if $C_j^\circ(\bm{\omega}^\prime) = \emptyset$, then $G_\circ[B_j^\circ(\bm{\omega}^\prime)]$ is empty or complete,   
\item if $C_j^\circ(\bm{\omega}^\prime) \neq \emptyset$, then $G_\circ[\{k\} \cup C_j^\circ(\bm{\omega}^\prime)]$ is connected for all $k \in B_j^\circ(\bm{\omega}^\prime)$. 
\end{itemize}
Hence, by the condition of $U$, for each $j \in W$,  
\begin{itemize}
\item $G[C_j(\bm{\omega}^\prime)]$ is empty or connected,  
\item if $C_j(\bm{\omega}^\prime) = \emptyset$, then $G[\bigcup_{S \in \mathcal{S}} S]$ is empty or connected for every $\mathcal{S} \subseteq \mathcal{B}_j(\bm{\omega}^\prime)$,  
\item if $C_j(\bm{\omega}^\prime) \neq \emptyset$, then $G[S \cup C_j(\bm{\omega}^\prime)]$ is connected for all $S \in \mathcal{B}_j(\bm{\omega}^\prime)$. 
\end{itemize}
Therefore, by \Cref{feasible_set}, we have $\widetilde{\mathcal{B}}_\circ^+ \subseteq \widetilde{\mathcal{B}}_\circ$. 
Consequently, by \Cref{feasible_set_core}, it follows that $\widetilde{\mathcal{B}}_\circ = \widetilde{\mathcal{B}}_\circ^+$.

\section{Connection to Risk Minimization} 
\label{background}
Let $\bm{Y}$ denote the random response vector at these input locations. 
The prediction vector $\bm{f}$ is assumed to serve as a proxy for the underlying mean vector $\mathbb{E}[\bm{Y}]$, where $\mathbb{E}$ denotes expectation with respect to the underlying probability measure of $\bm{Y}$. 
In this setting, we seek a vector $\bm{g} \in \mathbb{R}^n$ that is piecewise constant over contiguous regions and minimizes the squared prediction risk $\mathbb{E}\bigl[\|\bm{Y} - \bm{g}\|_2^2\bigr]$. 
For any fixed $\bm{g}$, the prediction risk admits the decomposition 
\begin{align} 
\label{prediction_error} 
\mathbb{E}\bigl[\|\bm{Y} - \bm{g}\|_2^2\bigr] = \mathbb{E}\bigl[\|\bm{Y} - \mathbb{E}[\bm{Y}]\|_2^2\bigr] + \|\mathbb{E}[\bm{Y}] - \bm{g}\|_2^2. 
\end{align} 
Since the first term of \Cref{prediction_error} is independent of $\bm{g}$, this risk minimization is equivalent to minimizing $\|\mathbb{E}[\bm{Y}] - \bm{g}\|_2^2$. 
Because the underlying mean vector $\mathbb{E}[\bm{Y}]$ is usually unknown, we approximate it by the prediction vector $\bm{f}$. 
Thus, the surrogate partition is obtained by solving $\mathrm{GPP}(\Omega_0, \bm{f})$, which identifies an assignment vector $\bm{\omega} \in \Omega_0$ that minimizes $\|\bm{f} - g(\bm{\omega}; \bm{f})\|_2^2$.

\section{MILP Formulation for Obtaining $\widetilde{B}_\circ$} 
\label{MILP} 
Let $\mathcal{N}_{jk}^\circ \coloneqq \{a \in V_\circ : \{k, a\} \in E_\circ(V), a \in C_j^\circ(\underline{\omega})\}$. 
Let $\mathcal{A}_j$ be the set of directed edges obtained from the edges of $G_\circ[C_j^\circ(\underline{\omega})]$. 
We introduce the following decision variables:
\begin{itemize}
\item $z_k \in \{0, 1\}$ indicates whether the $k$-th sublabel group is a boundary group, with $z_k=1$ if and only if it is a boundary group. 
\item $e_j \in \{0, 1\}$ indicates whether all groups in $C_j^\circ(\underline{\omega})$ are selected, with $e_j=1$ if and only if all such groups are selected. 
\item $r_{jk} \in \{0, 1\}$ indicates the root, $s_{jk} \geq 0$ denotes the flow supplied by the root, and $h_{jab} \geq 0$ denotes the flow along $(a,b) \in \mathcal{A}_j$. 
\item $u_{jk} \in \{0, 1\}$ indicates whether the $k$-th sublabel group is adjacent to at least one unselected group in $C_j^\circ(\underline{\omega})$. 
\item $w_k \in \{0, 1\}$ indicates whether a selected group is adjacent to fewer than two remaining core labels, with $w_k = 1$ if this condition holds. 
\item $v_{jk} \in \{0, 1\}$ indicates whether the $k$-th sublabel group is assigned to label $j$ whose core disappears completely. 
\end{itemize} 
We solve 
\begin{align} 
\max \sum_{k \in V_\circ} \sum_{i \in V : \sigma_i = k} \Bigl(f(\bm{x}_i) - |\{i^\prime \in V : \sigma_{i^\prime} = k\}|^{-1} \sum_{i^\prime \in V : \sigma_{i^\prime} = k} f(\bm{x}_{i^\prime})\Bigr)^2 z_k 
\end{align} 
subject to 
\begin{align}
\label{core1} 
z_k = 0 \quad (\forall k \in V_\circ \text{ satisfying } |S_k| < 2), \\   
\label{core2} 
e_j \leq z_k \quad (\forall j \in W, \forall k \in C_j^\circ(\underline{\omega})), \\ 
\label{core3} 
e_j \geq 1 - |C_j^\circ(\underline{\omega})| + \sum_{k \in C_j^\circ(\underline{\omega})} z_k \quad (\forall j \in W), \\  
\label{core4} 
\sum_{k \in C_j^\circ(\underline{\omega})} r_{jk} = 1 - e_j \quad (\forall j \in W), \\ 
\label{core5} 
\sum_{k \in C_j^\circ(\underline{\omega})} s_{jk} = \sum_{k \in C_j^\circ(\underline{\omega})}(1 - z_k) \quad (\forall j \in W), \\ 
\label{core6} 
r_{jk} \leq 1 - z_k \quad (\forall j \in W, \forall k \in C_j^\circ(\underline{\omega})), \\
\label{core7} 
s_{jk} \leq |C_j^\circ(\underline{\omega})| r_{jk} \quad (\forall j \in W, \forall k \in C_j^\circ(\underline{\omega})), \\ 
\label{core8} 
s_{jk} + \sum_{a : (a, k) \in \mathcal{A}_j} h_{jak} - \sum_{b : (k,b) \in \mathcal{A}_j} h_{jkb} = 1 - z_k \quad (\forall j \in W, \forall k \in C_j^\circ(\underline{\omega})), \\ 
\label{core9} 
h_{jab} \leq |C_j^\circ(\underline{\omega})| (1 - z_a) \quad (\forall j \in W, \forall (a, b) \in \mathcal{A}_j), \\ 
\label{core10} 
h_{jab} \leq |C_j^\circ(\underline{\omega})| (1 - z_b) \quad (\forall j \in W, \forall (a, b) \in \mathcal{A}_j), \\  
\label{boundary1} 
u_{jk} \geq 1 - z_a \quad (\forall j \in W, \forall k \in V_\circ \text{ satisfying } \mathcal{N}_{jk}^\circ \neq \emptyset, \forall a \in \mathcal{N}_{jk}^\circ), \\ 
\label{boundary2} 
u_{jk} \leq \sum_{a \in \mathcal{N}_{jk}^\circ} (1 - z_a) \quad (\forall j \in W, \forall k \in V_\circ \text{ satisfying } \mathcal{N}_{jk}^\circ \neq \emptyset), \\ 
\label{boundary3} 
u_{jk} = 0 \quad (\forall j \in W, \forall k \in V_\circ \text{ satisfying } \mathcal{N}_{jk}^\circ = \emptyset), \\ 
\label{boundary4} 
w_k \leq z_k \quad (\forall k \in V_\circ), \\ 
\label{boundary5} 
\sum_{j \in W} u_{jk} \leq 1 + m (1 - w_k) \quad (\forall k \in V_\circ), \\ 
\label{boundary6} 
\sum_{j \in W} u_{jk} \geq 2 (z_k - w_k) \quad (\forall k \in V_\circ), \\  
\label{boundary7} 
2 w_k \leq \sum_{j \in W} u_{jk} + \sum_{j \in W} v_{jk} \quad (\forall k \in V_\circ), \\  
\label{nocore1} 
v_{jk} \leq e_j \quad (\forall j \in W, \forall k \in V_\circ), \\ 
\label{nocore2} 
v_{jk} + v_{jk^\prime} \leq 1 \quad (\forall j \in W, \forall k, k^\prime \in V_\circ \text{ satisfying } k \neq k^\prime \text{ and } \{k, k^\prime\} \notin E_\circ(V)). 
\end{align}
\Cref{core1,core2,core3,core4,core5,core6,core7,core8,core9,core10} ensure that, for each label, the unselected core groups are empty or connected, using the flow variables when at least one core group remains. 
\Cref{boundary1,boundary2,boundary3,boundary4,boundary5,boundary6,boundary7,nocore1,nocore2} characterize the admissible connections of the selected boundary groups to the labels. 
Let $z_k^*$ denote the optimal value of $z_k$ in this problem. 
Then 
\begin{align}
\widetilde{B}_\circ = \{k \in V_\circ : z_k^* = 1\}. 
\end{align}

\section{Experimental Details} 
\subsection{Data and Software Resources}  
\label{source} 
\Cref{source_table} summarizes the data and software resources used in our experiments. 

\subsection{Illustration of Bridge-like Connections} 
\label{bridge_like_connections} 
\begin{figure*}[p] 
\begin{center}
\includegraphics[width=0.9\linewidth]{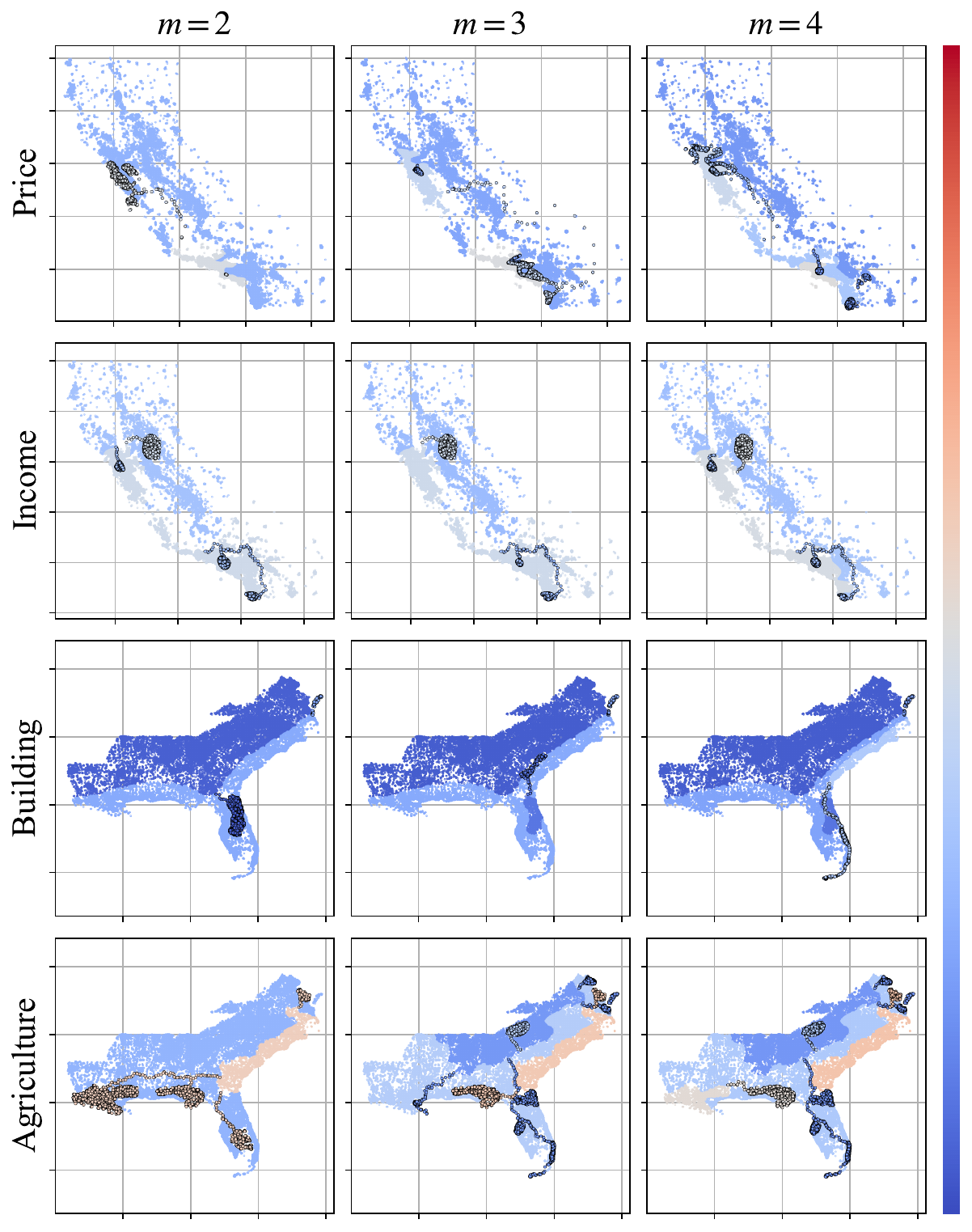}  
\end{center} 
\caption{
Greedily refined feasible partitions in $\Omega_0$ with $l = 30$. 
For each target variable, we show the trial with the largest value of $\|\bm{f} - g([1,\ldots,1]^\top; \bm{f})\|_2$. 
The panels show the cluster-wise projection $g(\bm{\omega}; \bm{f})$ for the resulting partitions. 
Vertices whose assignments differ from those obtained by our approach are outlined in black after matching the cluster labels to minimize the number of differing assignments. 
Colors indicate the levels of the projected values and are scaled separately for each target variable. 
}
\label{performance_bridge_30}
\end{figure*}

\begin{figure*}[p] 
\begin{center}
\includegraphics[width=0.9\linewidth]{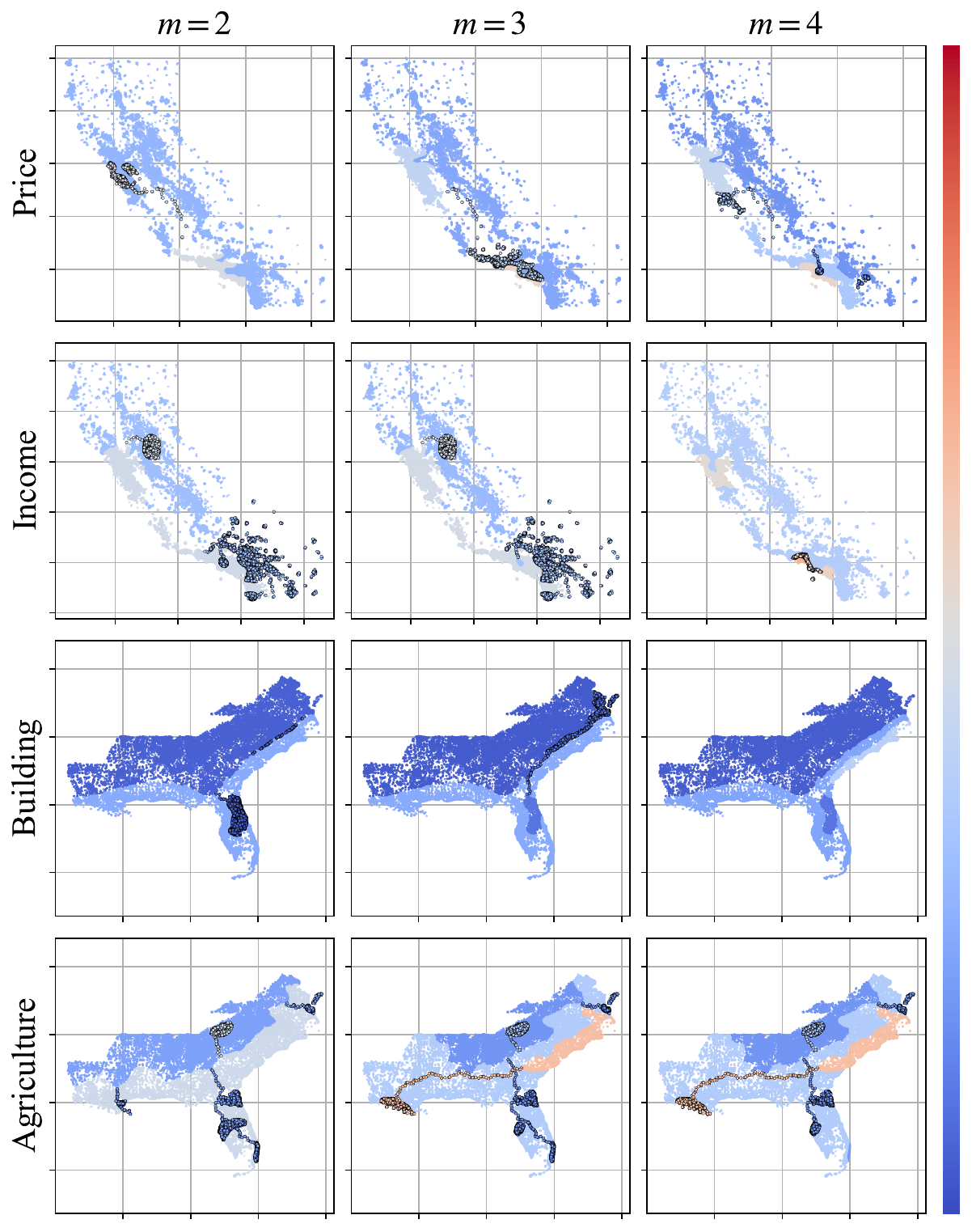}  
\end{center} 
\caption{
Greedily refined feasible partitions in $\Omega_0$ with $l = 15$. 
The figure follows the same layout as \Cref{performance_bridge_30}. 
}
\label{performance_bridge_15}
\end{figure*}

Feasible partitions in $\Omega_0$ are constructed by greedily relabeling the partitions obtained by our approach. 
At each iteration, we consider a sublabel group and a target label, relabel all vertices in the group and along a shortest path connecting the group to the target-label region, and select the feasible relabeling that yields the largest decrease in the objective. 
The procedure terminates when no feasible relabeling decreases the objective. 
Although this procedure further decreases the objective, as shown in \Cref{performance1_omega0}, the resulting partitions contain bridge-like connections, as shown in \Cref{performance_bridge_30,performance_bridge_15}. 

\subsection{Illustration of Boundary Adjustment}
\label{boundary_demo}
\begin{figure*}[p] 
\begin{center}
\includegraphics[width=0.9\linewidth]{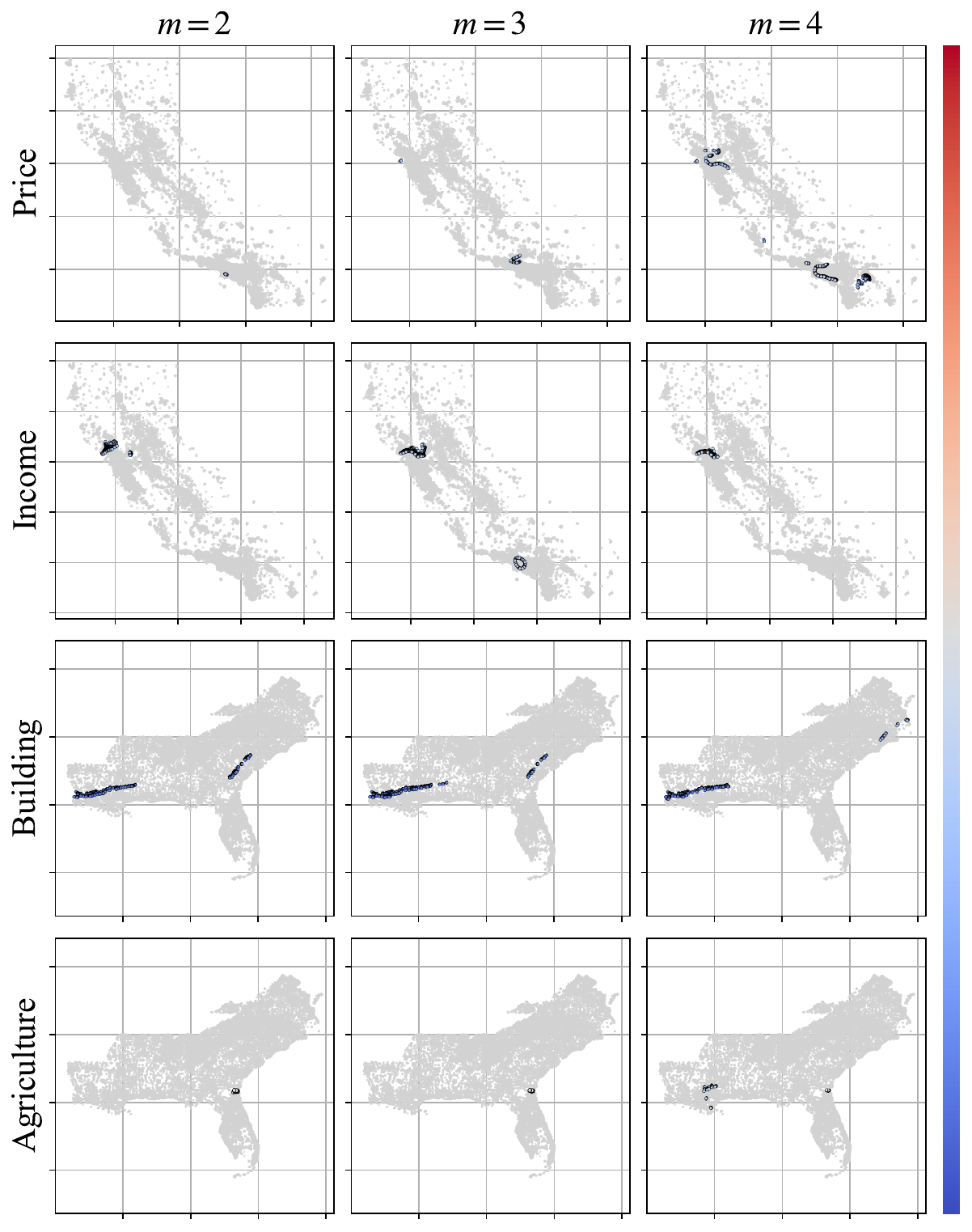}  
\end{center} 
\caption{
Differences between our approach and MIQCP for $l = 30$. 
For each target variable, we show the trial with the largest value of $\|\bm{f} - g([1,\ldots,1]^\top; \bm{f})\|_2$. 
The panels highlight the vertices whose assignments differ between the two methods after matching the cluster labels to minimize the number of differing assignments. 
Highlighted vertices are outlined in black and colored according to their prediction values, while the remaining vertices are shown in gray. 
Colors are scaled separately for each target variable. 
}
\label{performance2_diff}
\end{figure*}

\begin{figure*}[p] 
\begin{center}
\includegraphics[width=0.9\linewidth]{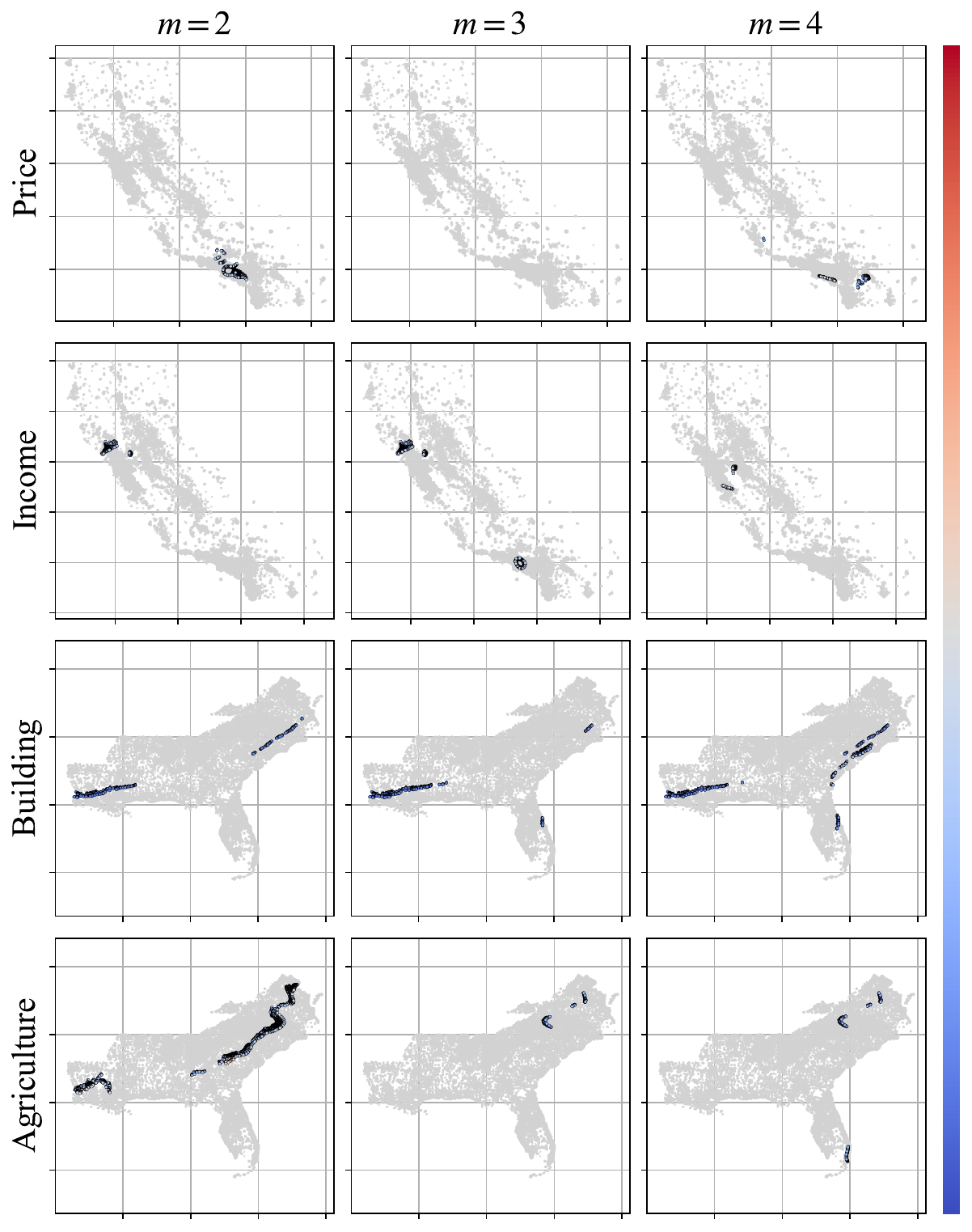}  
\end{center} 
\caption{
Differences between our approach and MIQCP for $l = 15$. 
The figure follows the same layout as \Cref{performance2_diff}. 
}
\label{performance_15_diff}
\end{figure*}

\Cref{performance2_diff,performance_15_diff} illustrate the changes in the partitions resulting from the boundary adjustment applied to the MIQCP solutions. 

\begin{table*}[h]
    \centering
    \caption{
    Data and software resources. 
    }
    \begin{threeparttable}[]
    \begin{tabular}{l >{\raggedright\arraybackslash}p{8.5cm} l} 
        \toprule
        Name & URL & License \\  
        \midrule 
        California Housing 
        & \url{https://scikit-learn.org/stable/api/sklearn.datasets.html}
        & Not specified \\ 
        National Risk Index 
        & \url{https://www.fema.gov/flood-maps/products-tools/national-risk-index}
        & U.S. Government Work \\ 
        Census Tracts 
        & \url{https://www.census.gov/cgi-bin/geo/shapefiles/index.php}
        & U.S. Government Work \\ 
        GPy 
        & \url{https://gpy.readthedocs.io/en/deploy/}
        & BSD 3-Clause \\ 
        scikit-learn 
        & \url{https://scikit-learn.org/}
        & BSD 3-Clause \\  
        SciPy 
        & \url{https://scipy.org/}
        & BSD 3-Clause \\ 
        SCIP 
        & \url{https://www.scipopt.org/}
        & Apache 2.0 \\ 
        PySCIPOpt 
        & \url{https://pyscipopt.readthedocs.io/}
        & MIT \\ 
        \bottomrule 
    \end{tabular} 
    \end{threeparttable}
\label{source_table}
\end{table*}

\begin{table*}[h]
    \centering
    \caption{
    Comparison of partitions in $\Omega$ and $\Omega_0$. 
    The mean and standard deviation of the error over $30$ trials are reported. 
    The definition of the error is the same as in \Cref{performance1}. 
    }
    \begin{threeparttable}[]
    \begin{tabular}{cccccc} 
    \toprule
    \multirow{2.5}{*}{Target} & \multirow{2.5}{*}{$m$} & \multicolumn{2}{c}{Partition in $\Omega$} & \multicolumn{2}{c}{Partition in $\Omega_0$} \\ 	
    \cmidrule(lr){3-4} \cmidrule(lr){5-6} & & $l = 15$ & $l = 30$ & $l = 15$ & $l = 30$ \\ 
    \midrule
    \multirow{3}{*}{Price} & 2 & $77.3~(9.1)$ & $76.5~(9.4)$ & $72.3~(9.3)$ & $71.1~(9.2)$ \\
     & 3 & $64.6~(5.3)$ & $63.9~(4.9)$ & $61.9~(5.0)$ & $60.6~(4.8)$ \\
     & 4 & $56.4~(3.7)$ & $55.4~(4.0)$ & $52.2~(3.9)$ & $50.2~(5.0)$ \\
    \midrule
    \multirow{3}{*}{Income} & 2 & $87.1~(2.6)$ & $86.5~(3.0)$ & $74.5~(7.9)$ & $72.8~(8.7)$ \\
     & 3 & $75.0~(5.3)$ & $73.8~(5.4)$ & $64.5~(5.4)$ & $63.6~(4.4)$ \\
     & 4 & $63.8~(3.7)$ & $63.0~(3.0)$ & $60.5~(4.6)$ & $58.7~(4.1)$ \\
    \midrule
    \multirow{3}{*}{Building} & 2 & $59.8~(2.0)$ & $58.4~(0.8)$ & $57.1~(4.9)$ & $54.2~(4.4)$ \\
     & 3 & $47.0~(3.3)$ & $45.9~(2.8)$ & $45.1~(3.8)$ & $42.9~(3.1)$ \\
     & 4 & $39.4~(2.0)$ & $38.3~(1.9)$ & $37.6~(1.6)$ & $36.7~(1.7)$ \\
    \midrule
    \multirow{3}{*}{Agriculture} & 2 & $82.5~(1.2)$ & $82.4~(0.7)$ & $70.1~(5.8)$ & $67.5~(5.1)$ \\
     & 3 & $72.0~(2.5)$ & $70.9~(2.1)$ & $61.5~(4.1)$ & $57.3~(3.5)$ \\
     & 4 & $66.6~(1.8)$ & $65.7~(1.7)$ & $57.7~(4.0)$ & $54.5~(2.8)$ \\
    \bottomrule 
    \end{tabular}
    \end{threeparttable} 
    \label{performance1_omega0}
\end{table*} 

\begin{figure*}[p] 
\begin{center}
\includegraphics[width=\linewidth]{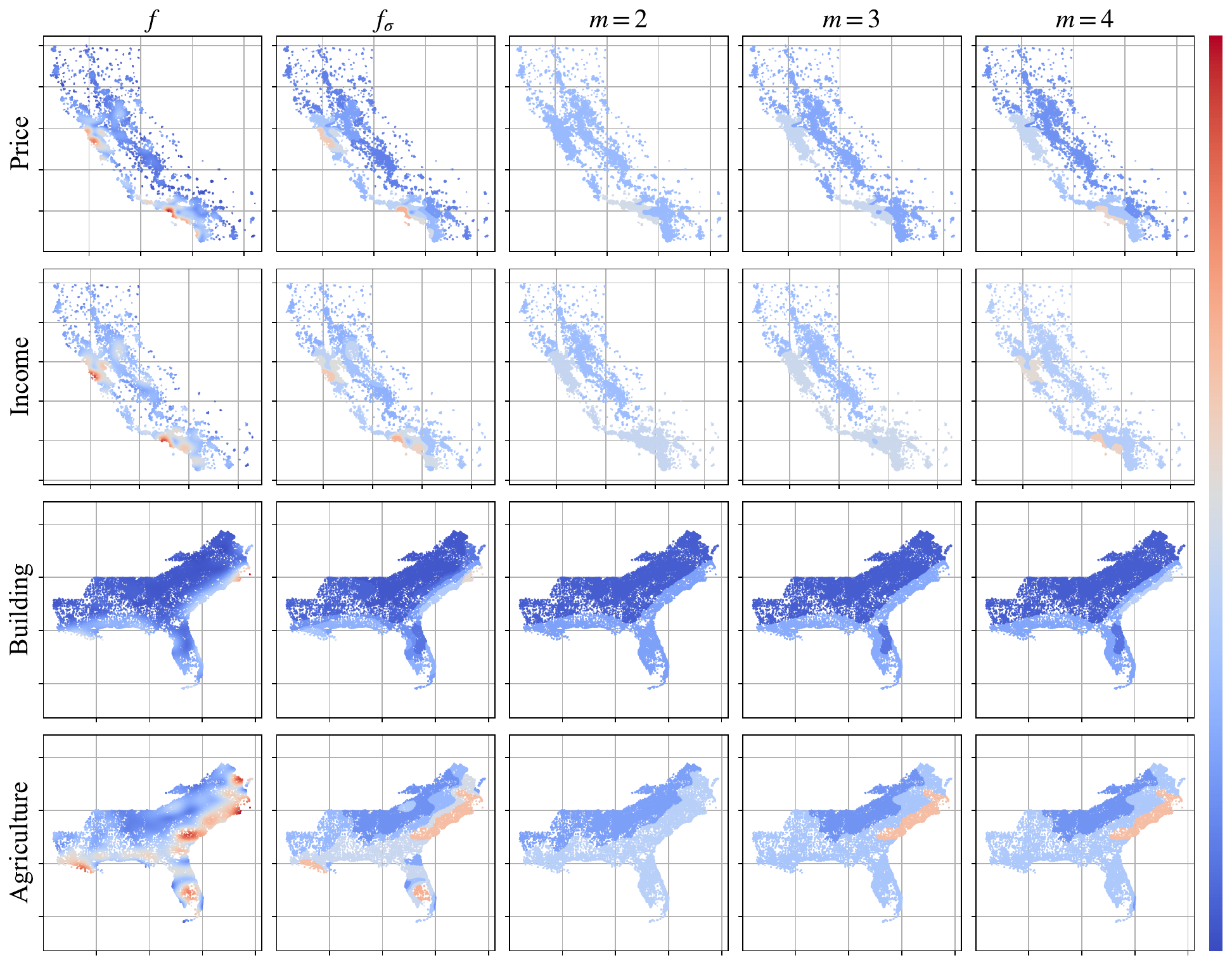}  
\end{center} 
\caption{
Spatial partitions produced by our approach with $l = 15$. 
The figure follows the same layout as \Cref{performance2}. 
}
\label{performance_15}
\end{figure*} 

\end{document}